\documentclass{article}

\PassOptionsToPackage{numbers}{natbib}

\usepackage[preprint]{neurips_2021}
\newif\ifaddchecklist

\usepackage[utf8]{inputenc} 
\usepackage[T1]{fontenc}    
\usepackage{hyperref}       
\usepackage{url}            
\usepackage{booktabs}       
\usepackage{amsfonts}       
\usepackage{nicefrac}       
\usepackage{microtype}      
\usepackage{xcolor}         
\usepackage{amssymb}
\usepackage{graphicx}
\usepackage{subcaption}
\usepackage{amsmath}
\usepackage{comment}
\usepackage{enumitem}

\usepackage{bbm}
\usepackage{amsthm}
\usepackage{float}
\usepackage{mathrsfs, pifont}
\usepackage{mathbbol}
\usepackage{upgreek}
 \usepackage{mdframed}
\newmdenv[
  topline=false,
  bottomline=false,
  rightline=false,
  skipabove=\topsep,
  skipbelow=\topsep
]{siderules}

\DeclareMathOperator*{\argmin}{arg\,min}
\DeclareMathAlphabet{\mathpzc}{OT1}{pzc}{m}{it}
\usepackage{float}
\usepackage{mathrsfs, pifont}
\usepackage{mathbbol, bm}
\usepackage{tikz}

\DeclareMathAlphabet{\mathpzc}{OT1}{pzc}{m}{it}

\title{Understanding the Logit Distributions of Adversarially-Trained Deep Neural Networks}

\author{%
  Landan Seguin\thanks{Work done at Intel Labs}\\
  \texttt{landanjs@gmail.com} \\
  \And
  Anthony Ndirango\\
  Intel Labs\\
  \texttt{anthony.ndirango@intel.com}\\
  \And
  Neeli Mishra\\
  Columbia University\\
  \texttt{nm2786@columbia.edu}\\
  \And
  SueYeon Chung\\
  Columbia University\\
  \texttt{sc4485@columbia.edu}\\
  \And
  Tyler Lee\\
  Intel Labs\\
  \texttt{tyler.p.lee@intel.com}
}

\begin{document}

\maketitle

\begin{abstract}
Adversarial defenses train deep neural networks to be invariant to the input perturbations from adversarial attacks. Almost all defense strategies achieve this invariance through adversarial training i.e. training on inputs with adversarial perturbations. Although adversarial training is successful at mitigating adversarial attacks, the behavioral differences between adversarially-trained (AT) models and standard models are still poorly understood. Motivated by a recent study on learning robustness without input perturbations by distilling an AT model, we explore what is learned during adversarial training by analyzing the distribution of logits in AT models. We identify three logit characteristics essential to learning adversarial robustness. First, we provide a theoretical justification for the finding that adversarial training shrinks two important characteristics of the logit distribution: the max logit values and the "logit gaps" (difference between the logit max and next largest values) are on average lower for AT models. Second, we show that AT and standard models differ significantly on which samples are high or low confidence, then illustrate clear qualitative differences by visualizing samples with the largest confidence difference. Finally, we find learning information about incorrect classes to be essential to learning robustness by manipulating the non-max logit information during distillation and measuring the impact on the student's robustness. Our results indicate that learning some adversarial robustness without input perturbations requires a model to learn specific sample-wise confidences and incorrect class orderings that follow complex distributions.

\end{abstract}

\section{Introduction}

Deep neural networks (DNNs) have been shown to be sensitive to small, often imperceptible perturbations around input data \cite{intriguing}. This sensitivity has led to concerns deploying DNNs in security-critical systems such as autonomous cars, face recognition, and malware detection, since an adversary could significantly degrade the network's performance without direct access to the model \cite{practical-bb}. Though the sensitivity of deep neural networks to adversarial attacks remains poorly understood, it aligns with arguments that DNNs utilize input features that run counter to human intuition \cite{geirhos2018imagenet}. This may lead to unexpected behavior when encountering out-of-distribution data \cite{geirhos2020shortcut}. To counteract these adversarial perturbations, numerous defenses have been proposed, but only a few perform consistently against a wide range of attack algorithms\cite{obfuscated, adaptiveAttacks}. The most notable type of defense is adversarial training where adversarial examples are generated to maximize the loss, and the network's weights are updated to minimize the loss on the generated examples\cite{harnessing}. Although adversarial training is successful at mitigating adversarial attacks, it hurts accuracy on the original input data and adds significant complexity to the training procedure, limiting its adaption in practice \cite{adversarial-industry}. Additionally, how adversarial training changes the behavior of the network compared to standard training is still an open question. An answer to this could allow us to simplify the process of making networks more robust to attacks and even provide additional insight into when networks do and do not generalize beyond their training dataset.

In this paper, we analyze the logits, i.e. the inputs to the softmax activation function, of adversarial trained (AT) models. As demonstrated in \cite{ard}, a significant amount of robustness can be learned by distilling an AT model. This indicates that the softmax outputs carry significant information pertaining to adversarial robustness and seems to suggest that properly constraining the output of a model could yield robust predictions. We focus on logits since we find they more readily distill adversarial robustness. Adversarial defenses based on logit regularization have been proposed before. Methods such as logit squeezing and label smoothing\cite{alp, warde2016, summers2019} have shown some promise, but several papers have claimed they are not robust without including some input perturbations \cite{lsq_lsm, jin2020rethinking, fu2020label}. Through a series of experiments and analyses, we have attempted to characterize the information present in the logits of adversarially trained models and found they are different in many complex ways from the logits produced by non-robust models. This suggests to us that it is unlikely that simple logit regularizers can be used to create robust models.

In this work, we reproduce several recent findings about the information present in an adversarially trained network's logits and expand on them in these three ways:
\begin{enumerate}
    \item Adversarially trained networks exhibit maximum logit values and "logit gaps" that are drawn from a distribution with a lower mean and a more positive skew than non-robust networks. We justify this with an analytical model and further show how these distributions change during training.
    \item Beyond a simple shift in distribution, adversarially trained networks give confident predictions to different inputs than standard, non-robust networks. Further, on a given input, AT networks and standard networks predict a different set of likely class labels. 
    \item Robustness can be distilled to a student network, but the targets provided by an adversarially trained teacher are finely-tuned. Slight manipulations of the distillation targets can degrade or even wholly remove the resulting student's robustness.
\end{enumerate}


\section{Related work}

 Since the initial study on adversarial examples \cite{intriguing}, increasingly strong attack algorithms have been demonstrated \cite{harnessing, moosavi2016deepfool, carlini2017} and a variety of defenses have been proposed \cite{harnessing, papernot2016distillation, song2018pixeldefend}. Although most defenses have been shown to be robust to particular attacks,  \cite{obfuscated, adaptiveAttacks} demonstrated that most defenses are not robust to a wide variety of attacks. One reliably consistent defense method involves training on adversarial examples generated by projected gradient descent (PGD) \cite{madry2018towards}. PGD builds on adversarial training \cite{harnessing} by applying iterative FGSM \cite{kurakin2017}, initializing the optimization at a random point within the allowed norm ball, and training exclusively on adversarial examples. Several improvements to PGD adversarial training have been proposed \cite{zhang2019theo, Wang2020Improving, zhang2020attacks}.

A separate, less conclusive defense approach is to constrain the neural network's outputs, either the logits or the softmax outputs. \cite{warde2016} suggested label smoothing \cite{Szegedy2016rethinking} as an alternative to the network distillation defense \cite{papernot2016distillation} and observed that it improved adversarial robustness. Several other studies \cite{fu2020label, lsq_lsm, jin2020rethinking} demonstrated that label smoothing obfuscates the gradients from adversarial example optimization and is only robust to a small set of attacks. Another defense proposed in \cite{alp} was "logit squeezing", penalizing large logit values by adding a regularization term to the loss. \cite{engstrom2018evaluating, mosbach2019logit} question the robustness of adversarial logit pairing and subsequently, the robustness of logit squeezing. \cite{lsq_lsm} argue that using only output regularizers such as logit squeezing and label smoothing are insufficient to learn robustness but claim using the output regularizers with inputs perturbed with Gaussian noise leads to substantial robustness. Beyond simple regularizers, \cite{ard} showed that adversarially trained networks can transfer most of their robustness through knowledge distillation on clean inputs only. They demonstrate that the student's robustness can match or exceed the teacher's when adversarial examples are used during distillation.

Knowledge distillation \cite{hinton2015distilling} proposes training a model, the student, to mimic the outputs of a pre-trained model, the teacher. In early works, the student was trained to match the teacher's temperature-scaled softmax output using a KL-divergence loss. Since then, numerous distillation losses have been proposed (see the survey\cite{Gou2021}). Typically, this setup is used to increase the accuracy of a "low capacity" student by training with a "high capacity" teacher, but it has been shown that self-distillation, where students have the same capacity as the teacher, also leads to improved performance \cite{furlanello2018born}. Several works \cite{furlanello2018born, tang2021understanding, phuong2019towards} attempt to identify the information in the teacher's output. So far, three main mechanisms have been suggested: (1) regularization through an effect similar to label smoothing, (2) sample re-weighting i.e. scaling the gradients of samples where the teacher is less certain, and (3) inter-class correlations referred to as "dark knowledge".

\section{Methods}
\label{sec:methods}

We trained ResNet-18 \cite{he2016deep} models for both our CIFAR-10 and CIFAR-100 \cite{Krizhevsky09learningmultiple} experiments. The full set of hyperparameters and sweeps we performed are described in detail in Appendix \ref{sec:additional-methods}. We implemented these experiments using PyTorch \cite{NEURIPS2019_9015} and the TorchVision ResNet18 implementation.

To learn adversarial robustness without input perturbations, we follow \cite{ard} and distill an adversarial trained model using only the clean training data. We focus on the self-distillation scenario where both the student and the teacher are ResNet-18. We find the AT teacher's epoch has significant impact to the student's robustness, so we used the checkpoint from epoch 40 for all experiments since this provided the best trade-off between natural and adversarial accuracy. Additionally, we find distilling the logits with the L1 loss leads to consistently better student robustness. Justification for this setup and additional details are in Appendix \ref{sec:additional-methods}.

Following previous knowledge distillation studies \cite{furlanello2018born, tang2021understanding}, we attempt to understand how robustness is distributed in the sample-wise logits of an AT model by fixing the top $k$ values and performing two manipulations for the remaining classes. The "permute" manipulation applies a uniform random permutation to the bottom $(N - k)$ classes where $N$ is the number of classes. This corrupts the inter-class correlation information from the bottom classes, but preserves the logit distribution. The "average" manipulation sets the bottom $(N - k)$ class values to the average of their values. A visualization of these manipulations is provided in Figure \ref{fig:top_k}.

We define adversarial accuracy throughout the paper as the model's accuracy after the first two attacks in the AutoAttack suite \cite{croce2020reliable}, namely auto-PGD with cross entropy (APGD-CE) and auto-PGD with the difference of logits ratio loss (APGD-DLR). We follow their attack procedure by running APGD-CE first, then use APGD-DLR on any samples correctly classified after the initial attack. We believe these attacks are sufficient to analyze how much robustness is transferred to a student. We verify for a few models that the adversarial accuracies are similar to running the full AutoAttack suite and the output-diversified initialization (ODI) attack \cite{tashiro2020diversity}.

\section{Results}
\label{sec:results}
\subsection{Dissecting logit distributional shifts}
\label{sec:teacher_logit}
 In this section, we describe our analysis of the relationship between a network's output logit distribution and its robustness to adversarial attacks. We trained ResNet-18 based neural networks using standard and adversarial training methods on two common image-based datasets (CIFAR10 and CIFAR100)\cite{Krizhevsky09learningmultiple}. All of our experiments were run with 3 random seeds and results are shown as mean $\pm$ standard deviation, where appropriate. Table \ref{table:teacher-acc} shows the natural and adversarial accuracy for our standard and adversarially trained models on CIFAR10 and CIFAR100. For clarity, we will often focus on results from the CIFAR10 dataset, but the results stated below apply to each dataset except where otherwise specified. Results for CIFAR-100 are included in Appendix \ref{sec:additional}. In this work, we find that standard and robust networks produce logits that differ in four important ways:
\begin{itemize}
    \item The maximum logit values from robust networks are drawn from a distribution with a low mean and positive skew, whereas those from standard networks are drawn from a more symmetric distribution with a much larger mean. 
    \item The "logit gap" (the difference between the maximum logit value and second largest value) is smaller in robust networks. A smaller logit gap is, counter-intuitively, \textit{less likely} to yield a robust response for a given example, but required to yield robust responses overall.
    \item Standard networks and robust networks disagree significantly on which examples they classify confidently. Robust networks favor images on plain backgrounds while standard networks favor a more diverse set of examples.
    \item Standard networks and robust networks disagree significantly on which set of class labels they find most likely for a given example.
    
\begin{table}[h]
  \caption{Accuracies on CIFAR-10 and CIFAR-100}
  \label{table:teacher-acc}
  \centering
  \begin{tabular}{lllll}
    \toprule                \\
    & \multicolumn{2}{c}{CIFAR-10} & \multicolumn{2}{c}{CIFAR-100}\\
    Model     & $A_{natural}$ & $A_{adversarial}$ & $A_{natural}$ & $A_{adversarial}$ \\
    \midrule
    ResNet18  & 94.5\% $\pm$ 0.20\%  & 0.00\% $\pm$ 0.00\% & 76.1\% $\pm$ 0.58\%  & 0.00\% $\pm$ 0.00\% \\
    AT ResNet18 (Epoch 40)  & 74.0\% $\pm$ 0.63\%  & 41.5\% $\pm$ 0.63\% & 49.1\% $\pm$ 0.85\%  & 19.3\% $\pm$ 0.48\%\\
    \bottomrule
  \end{tabular}
\end{table}

\end{itemize}
\begin{figure}[h]
    \centering
    \includegraphics{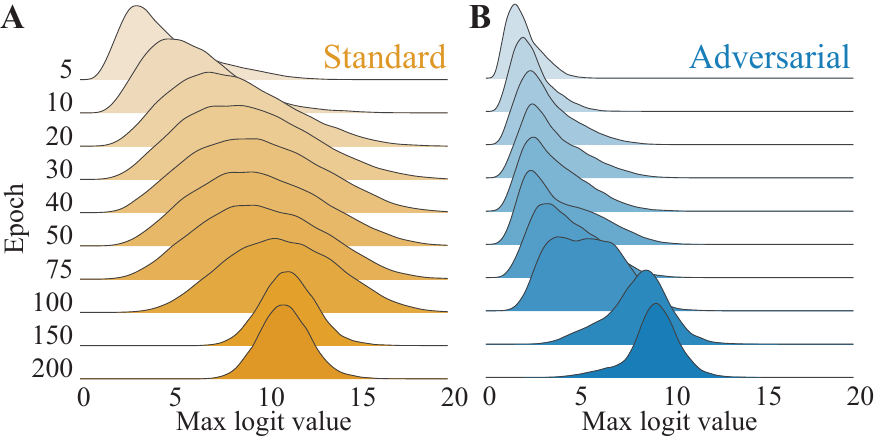}
    \caption{Evolution of the maximum logit values through 200 epochs of training. A) Standard training produces networks that quickly shift toward a symmetric, high mean distribution. B) Adversarial training produces networks that stay positively skewed with a low mean until very late in training.}
    \label{fig:logit_max}
\end{figure}

The maximum value of a network's logit outputs is often interpreted as proportional to the model's confidence in its response. As has been reported previously \cite{lsq_lsm}, we find that networks trained using adversarial training (AT) yield maximum logit values that are much smaller than those from networks trained with standard training methods (ST). However, the distribution is non-stationary, developing differently throughout each training method. Figure \ref{fig:logit_max} shows the evolution of the maximum logit distribution across many epochs for both ST and AT networks for the CIFAR10 training dataset. Throughout standard training with the cross entropy loss function, the maximum logit value first increases across all examples, resulting in a nearly symmetric distribution with a high mean value. In contrast, adversarial training more selectively increases a network's maximum logit value, producing a highly skewed distribution with a long positive tail throughout most of training. Interestingly, AT networks eventually converge to a largely symmetric, high mean distribution late in training while maintaining high robustness. This may relate to the observation that AT models exhibit significant overfitting after the learning rate decreases \cite{rice2020overfitting}. However, at convergence AT models continue to produce highly skewed maximum logit distributions on the test dataset. 

\begin{figure}[h]
    \centering
    \includegraphics{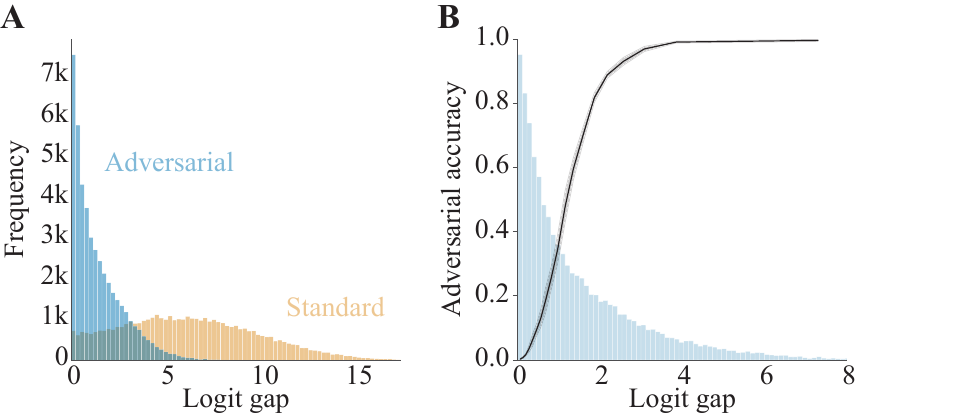}
    \caption{Comparing the logit gaps of adversarially trained (AT) models and standard trained (ST) models. (A) AT model logit gaps are significantly smaller than ST models with a peak at zero and few gaps larger than five. (B) The black line with shaded region represents the adversarial accuracy mean and standard deviation for three models across logit gaps. In contrast to ST models,  AT model's large logit gaps are a strong indicator of adversarial accuracy.}
    \label{fig:logit_gap}
\end{figure}
While the maximum logit value has long been suspected to play a key role in generalization and adversarial robustness (e.g. via label smoothing), more recent work has focused on the "logit gap" (the difference between the maximum logit value and the second largest value). The "logit gap" is intuitively strongly related to the ability of an adversarial attack to change the network's prediction, as it represents the distance that gradient-based perturbations must push the network's outputs. In addition to decreasing the maximum logit value, we find that AT also produces smaller logit gaps, with many near-zero values (Figure \ref{fig:logit_gap}A, also reported in \cite{lsq_lsm}). Throughout training, ST models more quickly develop a large, nearly symmetric logit gap distribution, while AT models retain a small, positively skewed logit gap distribution until near convergence (Figure \ref{fig:ridgeline-gap}). Despite this fact, the examples with low values are not themselves robust. Instead, these small values are a necessary condition for high-confidence, robust responses. This effect can also be seen in analytically tractable models as discussed in Appendix \ref{sec:analytical}. Figure \ref{fig:logit_gap}B shows the adversarial accuracy of an AT network on examples with increasing logit gaps.

To better understand the connection between adversarial robustness and the logit gap distribution, we developed an analytically solvable model that explains many of our empirical observations. Our analytical framework applies to any deep learning model and is especially suited to scenarios involving transfer learning or knowledge distillation where one has prior information provided by a teacher. We relegate the technical details of both the construction and the predictions of the solvable model to Appendix \ref{sec:analytical} and briefly summarize the highlights here.

Our analytical model first characterizes the space of logits leading to local minima of the cross-entropy loss conditioned on the distribution of the extrema of the logits. After identifying the space of feasible logit solutions, the model allows us to quantitatively study the effects that adversarial perturbations have on these optimal logits. In particular, we can calculate the change in the logit gap distribution induced by the adversarial perturbations. Our analysis leads us to predict that adversarial perturbations have the most pronounced negative effects on logits with high confidence on the training data. More precisely, logits with large logit gaps suffer larger perturbations to their response compared to logits with smaller gaps. Additionally, within the solvable model's framework, logits leading to almost perfect accuracy on the training data are more susceptible to adversarial perturbations relative to models which attain high, but not perfect, accuracy on the training data while maintaining small logit gaps. In Section \ref{sec:student_logit} we provide further evidence that these errors on the training set play a supportive role in enabling adversarial robustness. The analytical model further reveals that the presence of "small gapped" logits which make incorrect predictions on training samples serve to shield a model from large adversarial perturbations. All of these findings are either directly confirmed or strongly supported by our empirical results from experiments.
\begin{figure}[h]
    \centering
    \includegraphics{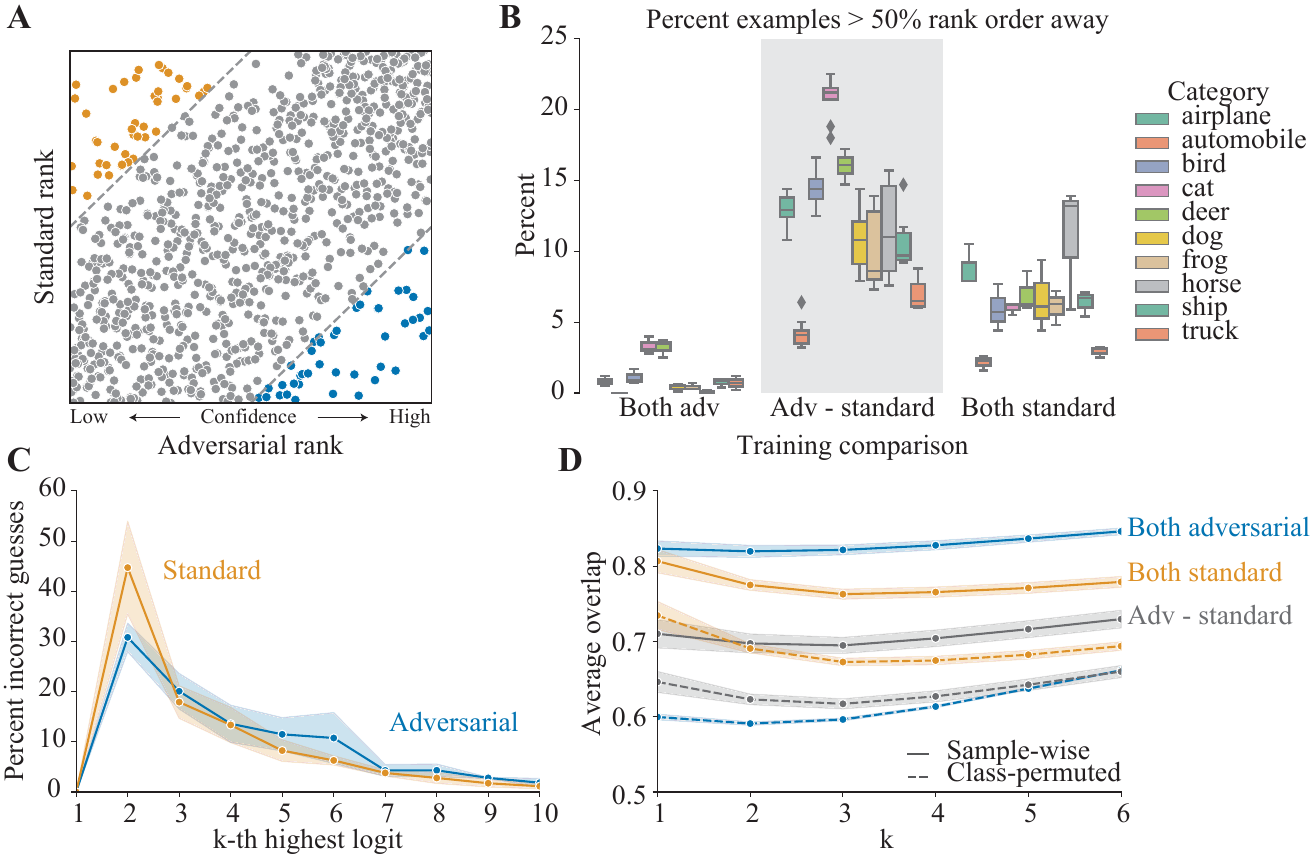}
    \caption{Predictions of adversarially trained (AT) and standard trained networks differ at the level of individual examples. A) Confidence values of both AT and ST networks were rank-ordered for every "frog" image in the CIFAR10 test dataset. The scatterplot shows large discrepencies between the rankings of individual images by the two types of networks. Images where the rank difference was greater than 50\% of the maximum value are shown in color. B) Comparison of the percent of examples with greater than 50\% rank order difference between individual networks. Three AT networks and three ST networks and their confidence levels on individual images from each of 10 classes were compared. AT networks rank images very similarly to each other but very differently than ST networks, where an average of 11.8\% of examples are ranked far apart. C) For both ST and AT networks, the average logit vector strongly predicts which classes the network will predict when it is \textit{incorrect}. ST networks often predict their second most probable class whereas AT slightly favor a broader set of classes. D) We used the "average overlap" statistic to compare the set of classes predicted on individual examples by ST and AT networks (solid lines). AT networks have high overlap with other AT networks but much lower overlap with ST networks, suggesting that AT networks are very similar but select different sets of likely classes than ST networks. This difference goes away if examples are compared with other images from the same class (dashed lines), suggesting that AT networks provide more diverse predictions to images from the same class. All comparisons are plotted as mean $\pm$ standard deviation.}
    \label{fig:logit_specifics}
\end{figure}

Beyond the distribution of logit values, ST and AT networks differ in their specific responses to individual examples. AT and ST networks are confident about different types of images, with AT networks favoring images on plain backgrounds and ST networks responding confidently to a more diverse set of images. These differences can be seen in Figure \ref{fig:confidence_images} in the appendix. This higher confidence in AT networks to objects on plain backgrounds is likely related to the recent demonstrations of shape preference within AT networks \cite{geirhos2018imagenet,li2021shapetexture}. To study this effect quantitatively, we rank ordered the confidence levels of each network on each image category. In this specific case, the rank ordering was performed on the softmax values rather than the logit values, as the normalization of the softmax makes the comparison more consistent across examples. Using the logit values produces qualitatively similar results, however. Figure \ref{fig:logit_specifics}A shows a comparison between the rank orderings of an AT and ST network for all "frog" images in the CIFAR10 test set. By looking at the fraction of images that had a rank difference of more than half of its maximum value, we found that an average of 11.8\% of examples across all classes were ranked very differently between AT and ST networks. AT networks were very consistent amongst themselves, however, differing greatly on only about 1.1\% of images.

Knowledge distillation research indicates that the non-max logit values relate to inter-class correlations present in the dataset \cite{tang2021understanding}. Thus, not only are the logit \textit{values} important, so too are the specific class indices for an individual example. For both ST and AT networks, these non-max logit values can be used to predict the errors made on natural and adversarial images alike. To see this, we first computed the average logit vector for each class across all correct answers. We then compared the fraction of incorrect guesses that corresponded to the k-th highest average logit value, for $k \in [1, 10]$ (Figure \ref{fig:logit_specifics}C). As this analysis suggests, those classes the model predicts are possible, though not the most likely, become its most likely errors when faced with an uncertain or perturbed example. ST and AT networks are quite similar in this regard. Given the utility of the rank-order of a model's logit outputs (i.e. higher rank order is more likely to become one of the model's incorrect guesses), we computed the average overlap \cite{webber2010similarity} between the logit rank orders for AT and ST networks. The average overlap at rank $k$ is computed as $AO@k = \frac{1}{k}\sum_{i=1}^{k} O(k)$, where $O(k) = \frac{1}{k}|L_{:k}^1 \cap L_{:k}^2|$, the proportion of intersection between the first $k$ indices in the rank order of two vectors. Different seeds of AT networks have significantly more overlap than AT and ST networks (Figure \ref{fig:logit_specifics}D, solid lines). Interestingly, this is driven by an increase in output variability for AT networks in response to examples of the same class, and not instead by a difference in the average logit output across examples. To see this, we permuted the examples from each class and computed the $AO@k$ again and found that AT networks showed less consistency than ST networks (Figure \ref{fig:logit_specifics}D, dashed lines). This suggests that AT networks produce logits that are more specifically tailored to the individual image than ST networks. In the next section we drill down on the importance of cross-class information in the AT model's logit outputs. We focus these analyses on the CIFAR10 dataset.

\subsection{Analyzing via logit distillation}
\label{sec:student_logit}
\begin{figure}[h]
    \centering
    \includegraphics{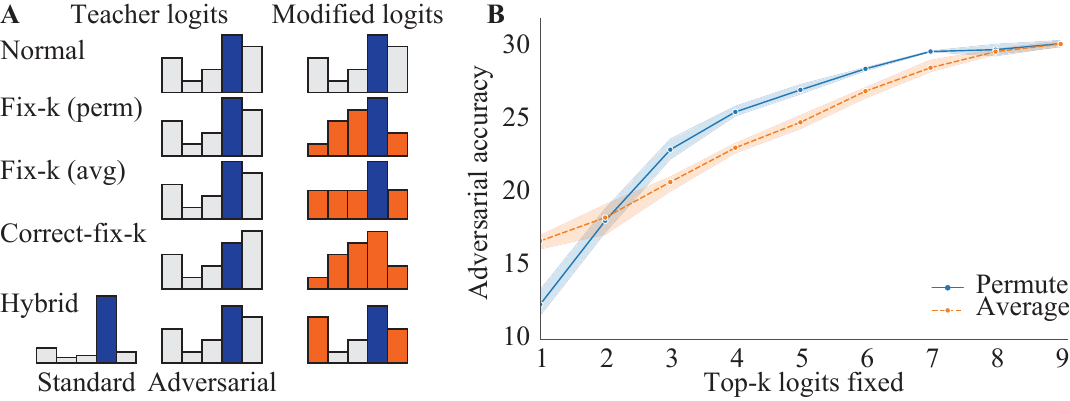}
    \caption{Fix the per sample top $k$ classes and permute the remaining classes. (A) A visual example of how the logit distribution changes after each type of manipulation where the blue bar indicates the ground truth class and the orange bars represent manipulated logit values (B) The top-3 values provide almost 80\% of the full robustness, but the top-7 values are required to be within 99\%. }
    \label{fig:top_k}
\end{figure}

To investigate how adversarial robustness is encoded in the cross-class information of model's logits, we used both AT and ST networks as teachers in the process of knowledge distillation. It has been recently demonstrated that robustness can be taught to a student network by training the student to match an AT teacher's softmax outputs alone \cite{ard}. This strongly suggests that the information required to ensure robust image representations is entirely contained within, and readily decoded from, the output distribution of an adversarially trained network. Thus, by manipulating the teacher outputs used to train a student network, we were able to readout important details about the way robust networks encode images. Figure \ref{fig:top_k}A describes the set of manipulations we performed. From these experiments, we found that:
\begin{itemize}
    \item Robustness is dependent on at least the top-7 logit values produced by the AT teacher
    \item A significant amount of cross-class information is encoded in the training examples that the AT teacher gets \textit{incorrect}.
    \item Hybrid models combining the AT and ST teachers produce students with remarkably little robustness.
    \item Student networks that are not robust show poor representations of adversarial images as early as the first residual block, similar to ST teachers
\end{itemize}

 AT teachers trained for up to 100 epochs showed a strong ability to distill their robustness. Interestingly, though teachers trained for longer retained high robustness, they could not distill it to their students. Based on our findings in the previous section, we believe this is due to the AT teachers overfitting on the training set and producing logit targets with a larger logit gap. Figure \ref{fig:ridgeline-gap} supports this interpretation, as the logit gaps on the test set remain small for these late checkpoint teachers, but the student networks only see the large gap training set outputs. Since the amount of robustness distilled to students varied by which teacher checkpoint was used, all experiments below were performed using the teacher checkpoint that produced maximum robustness (epoch 40).
Our first experiment attempted to characterize the number of CIFAR10 classes per example that carried relevant information in the logit vector. For each example in the training set, we fixed the top $k$ classes (by their logit values) and then either averaged or permuted the remainder. We then trained a student network using these modified teacher logit vectors and computed their adversarial robustness after convergence. We call this the "fix-k" experiment. The results of this can be found in Figure \ref{fig:top_k}B. For now we'll focus on the "permute" case, though the results for averaging are qualitatively similar. Keeping only the top 1 class provides the student with only the teacher's confidence and prediction for each training image. In this case, adversarial accuracy on the test dataset fell dramatically but remained at 12.43\%. While this accuracy is significantly lower than the student trained with unmodified teacher logits ("normal" distillation), we point out that it is far higher than a standard trained network, where we routinely measure 0 robust images using our suite of adversarial attacks. As we reduced the number of modified components of the teacher's logits, the students' adversarial accuracy increased, up until modifying all but the seven largest logit values for each example. That is, up to seven logit values per image carry relevant information about the AT teacher's robust response.

\begin{table}[h]
  \caption{Accuracies of students distilled using manipulated logits}
  \label{table:student-acc}
  \centering
  \begin{tabular}{lll}
    \toprule                \\
    Student     & $A_{natural}$ & $A_{adversarial}$ \\
    \midrule
    Normal & 75.2\% $\pm$ 0.30\%  & 30.4\% $\pm$ 0.34\% \\
    Fix-1 (perm)    & 77.6\% $\pm$ 0.6\% & 13.3\% $\pm$ 2.0\% \\
    Correct-fix-1 (perm)  & 92.1\% $\pm$ 0.19\% & 0.1\% $\pm$ 0.05\% \\
    Hybrid ST-AT  & 87.86\% $\pm$ 0.19\% & 0.00\% $\pm$ 0.00\% \\
    \bottomrule
  \end{tabular}
\end{table}

The "fix-k" experiment makes clear that robust predictions are encoded densely in AT networks' logit outputs. However, the retained adversarial accuracy in the student trained with "fix-1" modified logits suggests two possible explanations: 1) the robustness is due to the teacher providing a per-example confidence level to the student or 2) the robustness is due to the residual cross-class information provided by the teacher when it gets training examples wrong. To disambiguate these two options we performed a second modification to the "fix-1" logits - we corrected each incorrect teacher prediction from the training set (~19.6\% $\pm$ .6\% of examples) by swapping the logit values for the teacher's prediction and the ground truth label. As shown in Table \ref{table:student-acc}, this "correct-fix-1" experiment resulted in a student network with much higher natural accuracy but 0\% adversarial accuracy. Thus, it seems the retained adversarial accuracy provided by the max logit value from the teacher is primarily due to the information it contains about class relationships when it incorrectly predicts a training example.

In Figure \ref{fig:logit_specifics}D we showed that AT and ST networks predict significantly different sets of possible classes for individual images. It is unclear, though, whether that magnitude of an effect is critical in practice (though our "fix-k" experiments strongly suggest it). To investigate this, we created a hybrid set of teacher labels by merging the logit \textit{values} from the AT teacher with the \textit{class indices} from the ST teacher. If these labels could produce a robust student network, then we might be able to disentangle the challenge of creating logit vectors drawn from a specific distribution from the challenge of determining the appropriate cross-class relationships for each example. Student networks trained with these hybrid labels, however, were not robust to any examples in the CIFAR10 test dataset. 
\begin{figure}[h]
    \centering
    \includegraphics{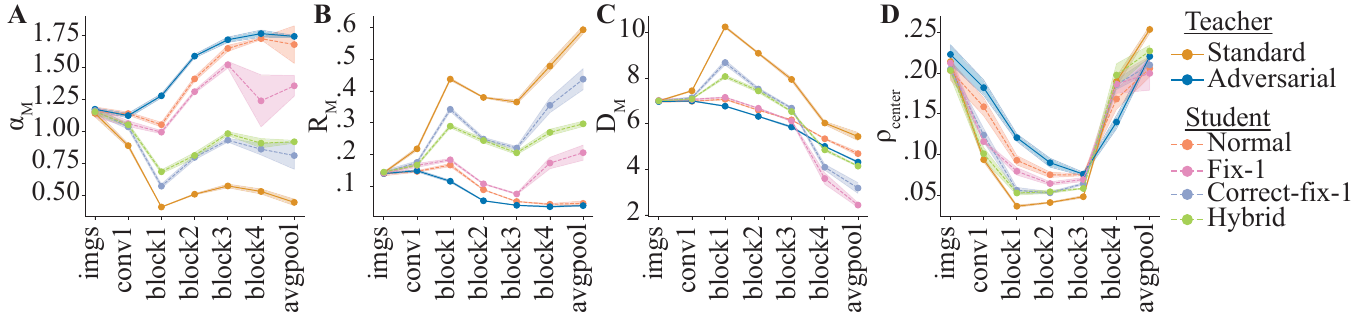}
    \caption{Manifold analysis of robust and non-robust teacher and student networks. The capacity for adversarially perturbed images (A) of non-robust student networks falls off in the early layers, similar to a standard trained teacher. This is largely due to an increase in manifold radius (B) amongst these networks. The manifold dimension (C) and center correlations (D) are similar across networks.}
    \label{fig:mftma}
\end{figure}
Adversarial accuracy provides a blunt measure of a network's robustness to adversarial perturbations, as it only measures differences at the network output and even then only whether or not a perturbation is strong enough to change the network's prediction. To provide further details about the scale of an adversarial image's perturbation throughout the network, we applied a recently developed manifold analysis technique \cite{chung2018classification, cohen2020separability} (see \cite{stephenson2019} for code used) to estimate the size and shape of the induced network representation as it propagates into the deeper layers. This technique is called replica mean field theoretic manifold analysis, or MFTMA, which formally connects the linear decodability of categorical data to its geometrical properties. While this analysis has primarily been applied to categorical data, we can use it to measure the separability of $\ell_{\infty}$ balls throughout a network. Given \textit{P} exemplar images, \textit{M} adversarial perturbations of each and \textit{N} dimensions, the MFTMA returns the manifold capacity, radius, dimension and center correlations. Manifold capacity is a measure of linear separability and refers to the maximum number of category manifolds that can be linearly separated given $N$ features. Manifold radius and dimension refers to the size and dimensionality of the category manifolds relevant for the linear classification. Center correlation measures the correlation between the center locations of the manifolds (refer to Appendix \ref{sec:additional-methods} for additional details). This analysis suggests that the radius of the manifold for images perturbed within an $\ell_{\infty}$ ball with radius $\epsilon = 8/255$ increases quickly in early layers of networks that are not adversarially robust (Figure \ref{fig:mftma}B). This analysis instead focuses on the "fix-1 (avg)" experiments. Student networks trained on the "corrected-fix-1" and the "hybrid" logits exhibited significantly decreased capacity for adversarially perturbed images, nearing that of the non-robust standard trained teacher network (Figure \ref{fig:mftma}A). Each tested network that exhibited non-zero adversarial accuracy showed increased capacity in the deeper layers, though the student trained with "fix-1" logits exhibited a final decrease in capacity in the last residual block.

\subsection{Promise of visual similarity}
Thus far, the results of our empirical analyses suggest that the outputs produced by AT networks are highly tuned and adversarial robustness falls off quickly if the information spread throughout the logit distribution is degraded. Here we note a promising lead on the origin of the cross-class information in the AT network's logits.  Starting with a seed image, we use cosine similarity on the logits to find a set of nearby images. Images nearby in the AT network's output space are very similar in both color and object shape (Figure \ref{fig:similarity}). In contrast, images nearby in the ST network's output space are diverse in appearance and primarily of the same class as the seed image. It seems intuitive that adversarial training would require that the network favor visual similarity, so this may prove to be a causal difference that results in the dense code that is evidenced in our "fix-k" experiments. We leave a more thorough analysis of this to future work.

\section{Discussion}
\label{sec:discussion}
Adversarial training remains the gold standard for producing models that are robust to adversarial attacks. A key goal of much recent research has been to better understand the effects of adversarial training on the network such that standard training methods can be made robust through simple modifications. Given the ability of a robust teacher to distill robustness to a student network using only its soft labels, we set out to better understand the structure in a robust network's logits in the hope that simple modifications of standard training could produce similar logits and thus ease robust distillation. \textit{Through a series of experiments and analyses, we have found that this seems unlikely}. 

Robust networks differ from nonrobust networks in many ways. Not only do their logit distributions show distinct features and evolve separately throughout training, but they process individual images very differently. Robust networks and nonrobust networks answer confidently on very different sets of images. They also differ significantly in how they distribute information across class indices within the logit vector. These differences appear critical to the development of robust responses, as adversarial accuracy in a student network falls after small manipulations to the logit vectors and can easily be wiped out entirely. This fall in adversarial accuracy occurs early in the network, suggesting some of these changes dramatically alter the network's initial training dynamics.

\paragraph{Limitations}
\label{sec:limits}
Given the consistency of our results through many experiments, multiple datasets and an analytical model, we are hopeful in the generality of our claims. However, we recognize that significant work remains to be done to demonstrate these effects across additional datasets, network architectures, training regimes and problem domains. We feel that a focused approach was necessary for this type of experimental exploration and hope to use this work as a foundation for broader research.

\paragraph{Broader Impact}
\label{sec:impact}
The challenge of adversarial attacks will continue as a growing concern as deep neural networks are deployed more broadly. For this, any work that is done to better understand how to mitigate these attacks could prove critical. Beyond the security risk that adversarial examples pose, they represent a significant gap in our understanding of how deep networks process the world. Safe and reliable deployment of DNNs depends critically on careful research into the failures of traditional training methods and the mechanisms through which alternative training methods offer improvements.

\clearpage
\small
\bibliography{references}
\bibliographystyle{unsrt}
\clearpage

\appendix
\renewcommand\thefigure{\thesection.\arabic{figure}}  
\section{Additional methods}
\label{sec:additional-methods}
\subsection{Computational requirements}
\label{sec:compute-req}
All experiments were run on an internal cluster. Individual experiments utilized computational units with 1 NVIDIA Titan Xp GPU, 12 CPU cores and 16 GB RAM each. On this hardware, the average runtime per training run are described in Table \ref{table:wall-time} In total, the results in this paper represent 136 total training runs.
\begin{table}[h]
  \caption{Experiment training time}
  \label{table:wall-time}
  \centering
  \begin{tabular}{lll}
    \toprule                \\
    & \multicolumn{2}{c}{Wall time} \\
    Training     & CIFAR10 & CIFAR100 \\
    \midrule
    Standard & 1.66 Hr  & 1.66 Hr\\
    Adversarial     & 18 Hr & 18 Hr\\
    Knowledge distillation     & 3.83 Hr & - \\
    \bottomrule
  \end{tabular}
\end{table}
Our experiments were developed using the PyTorch library (pytorch.org), which is open-sourced under the BSD license. Many of our experiments were run on the CIFAR10 and CIFAR100 datasets, which is publicly available under the MIT license.

\subsection{Experiment hyperparameters}

We use ResNet-18 \cite{he2016deep} models for our CIFAR-10 and CIFAR-100 \cite{Krizhevsky09learningmultiple} experiments. We train our models for 200 epochs using SGD with momentum of 0.9, learning rate of 0.1 decreased by 0.1 at epoch 100 and 150, batch size of 128, and weight decay of 2e-4. We apply random crops and horizontal flips as data augmentations. The architecture and hyperparameter were chosen to match \cite{ard} which we used as the basis for our ResNet18 implementation. For CIFAR-10, we removed 500 random training samples from each class to be used as a validation set. The validation set was used to check progress during training, but all reported results are from the provided CIFAR-10 test set. Results on CIFAR-100 were more sensitive to using a reduced training set, so CIFAR-100 results are from models trained on the full training set.

For adversarial training, we use the projected gradient descent (PGD) algorithm from \cite{madry2018towards} to generate the adversarial examples within an $\ell_{\infty}$ ball with radius $\epsilon = 8/255$. We apply 10 iterations of PGD with a step size of $2/255$. We use the same architectures and apply the same hyperparameters as our standard training. We use the PGD implementation from \cite{zhang2019theo}.

\subsection{Modifications to robust distillation}

\begin{figure}[h]
\begin{subfigure}{.5\textwidth}
    \centering
    \includegraphics[width=0.8\textwidth]{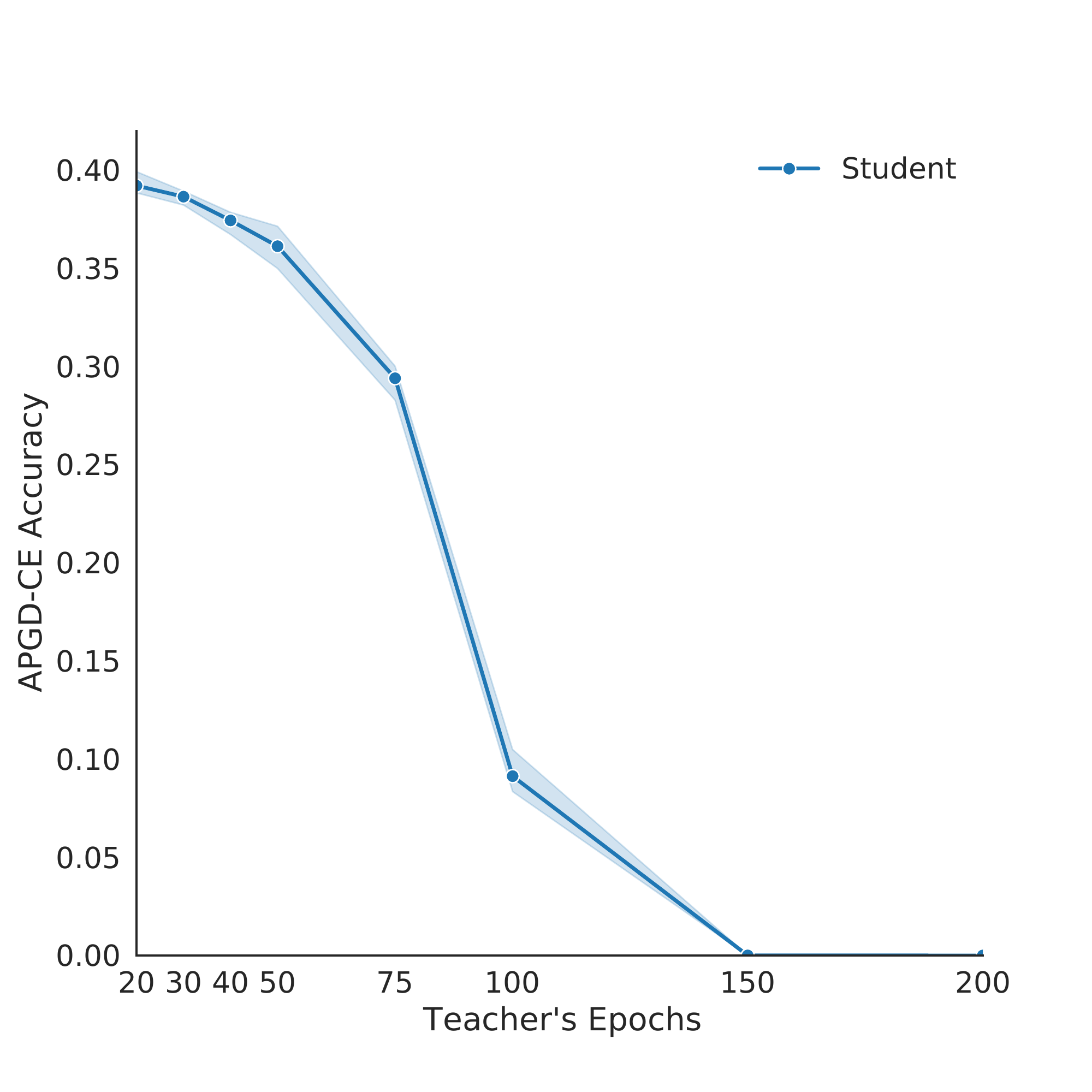}
    \caption{}
    \label{fig:adver_v_epoch}
\end{subfigure}
\begin{subfigure}{.5\textwidth}
    \centering
    \includegraphics[width=0.8\textwidth]{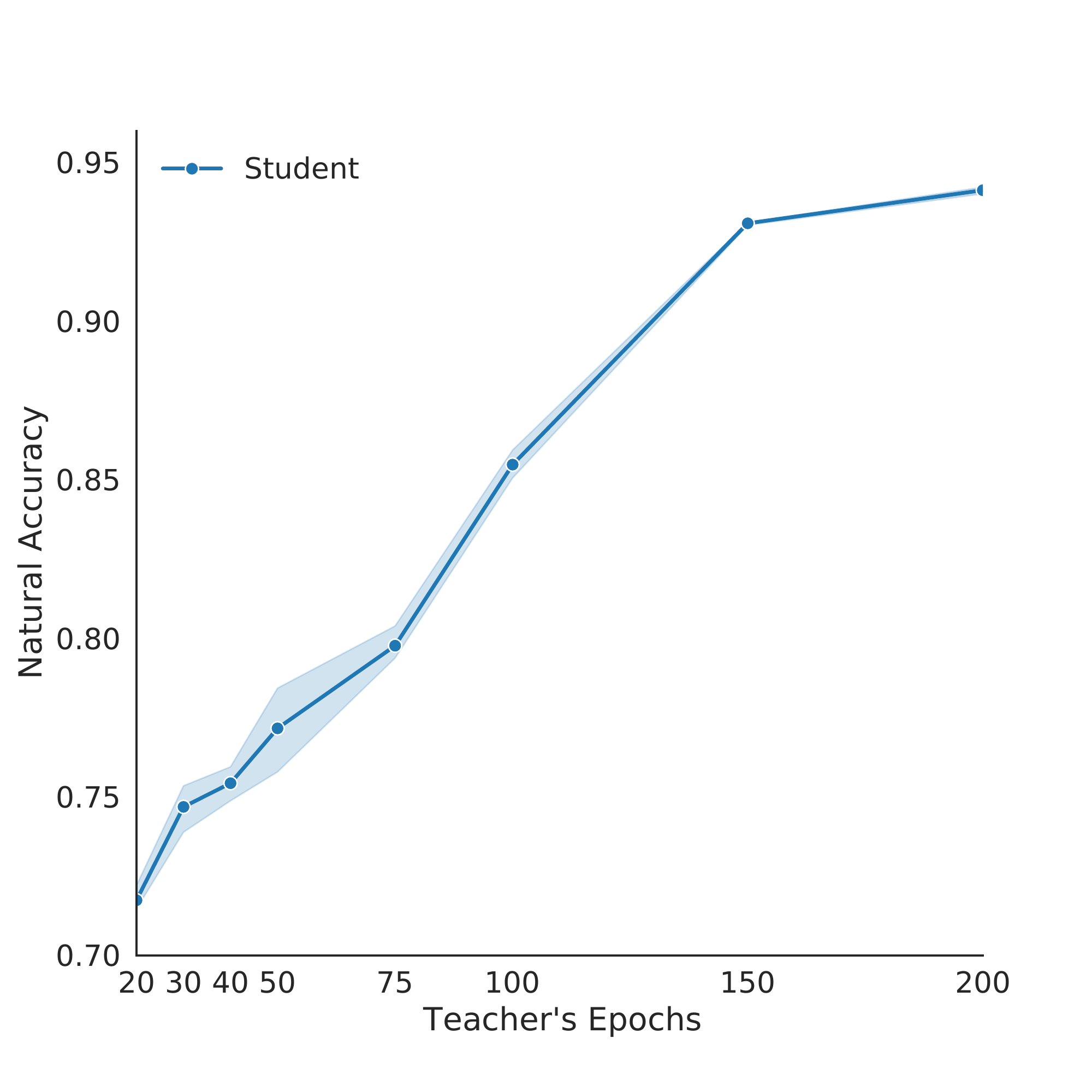}
    \caption{}
    \label{fig:nat_v_epoch}
\end{subfigure}
\caption{Student Accuracies vs Teacher's Epoch}
\label{fig:acc_v_epoch}
\end{figure}

\begin{figure}[h]
\begin{subfigure}{.5\textwidth}
    \centering
    \includegraphics[width=\textwidth]{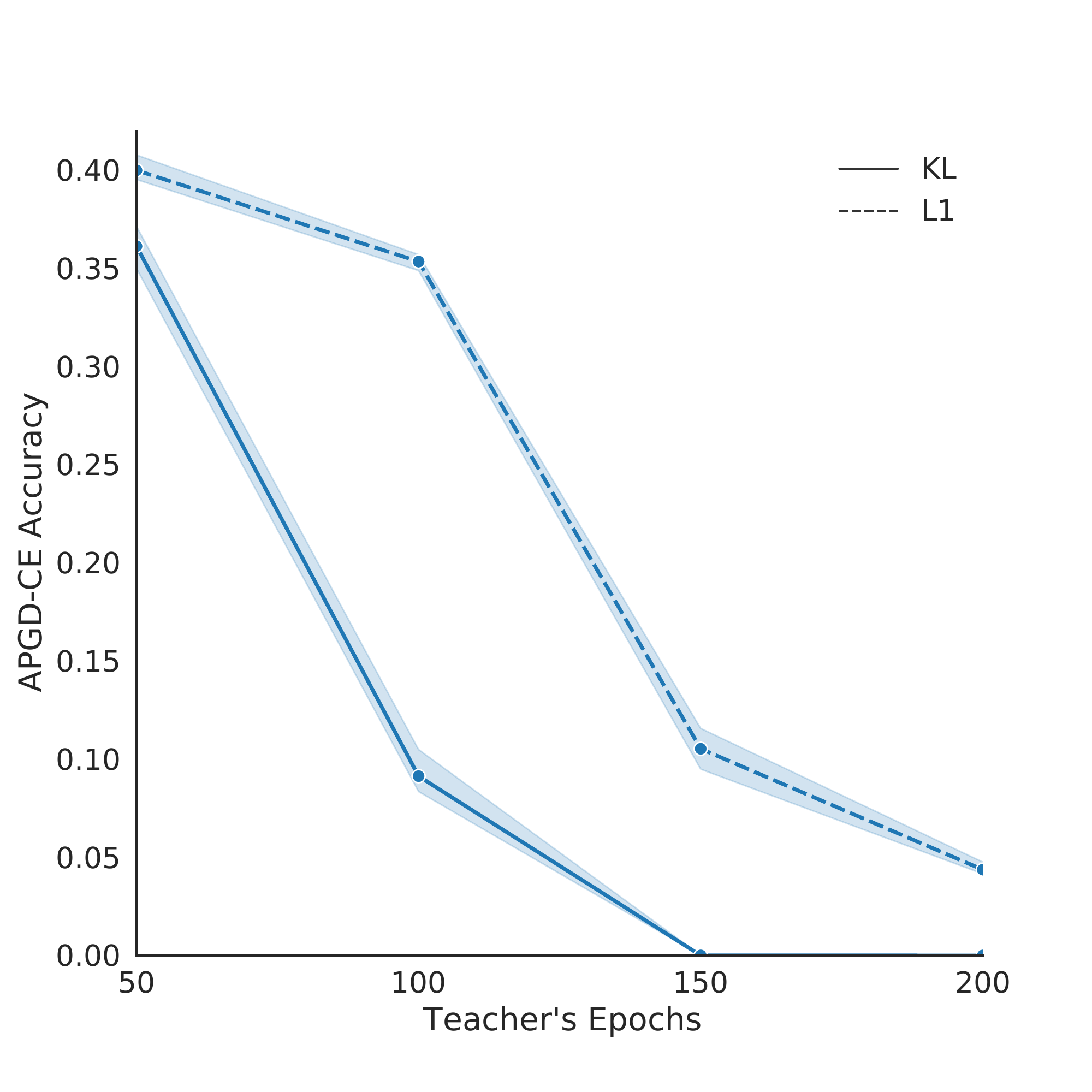}
    \caption{}
    \label{fig:adver_v_loss}
\end{subfigure}
\begin{subfigure}{.5\textwidth}
    \centering
    \includegraphics[width=\textwidth]{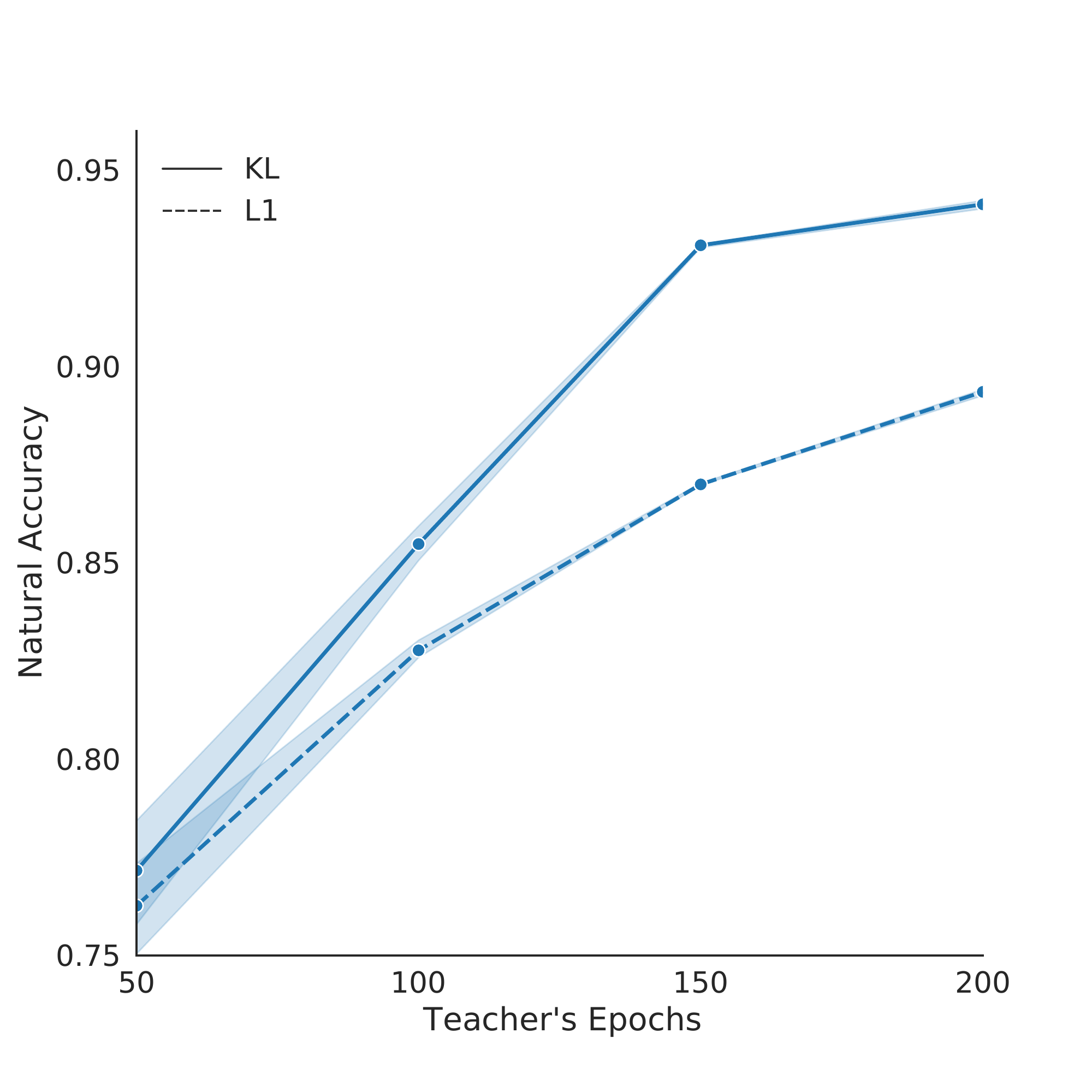}
    \caption{}
    \label{fig:nat_v_loss}
\end{subfigure}
\caption{Student Accuracies vs Loss}
\label{fig:acc_v_loss}
\end{figure}

Our work is focused on transferring adversarial robustness by distilling an adversarially trained model using only clean images. This method was demonstrated by \cite{ard}, but we make three observations and changes to the original setup:

\begin{enumerate}
    \item The teacher's number of training epochs has significant impact on the student's adversarial and natural accuracies. As shown in Figure \ref{fig:adver_v_epoch}, the student's adversarial accuracy ranges from nearly 40\% to 0\% when distilling from epoch 20 and 200 respectively. We believe this is related to the observation that adversarially trained models exhibit significant overfitting starting a few epochs after the first learning rate decay \cite{rice2020overfitting}. We use the teacher's epoch 40 checkpoint since it provides the best trade-off between natural and adversarial accuracy. 
    \item Distilling the teacher's logits using the L1 loss consistently transfers more robustness than distilling the teacher's softmax output using the KL divergence loss as shown in Figure \ref{fig:adver_v_loss}. In particular for epoch 100 teacher, there is a 25\% absolute difference between student's trained on logit versus softmax distillation. The improvement is likely similar to increasing the temperature as in \cite{ard}, but no temperature tuning is required. Additionally, logit distillation leads to lower L1 and KL loss values, suggesting the improvement is from the student getting closer to matching the teacher. 
    \item When training with knowledge distillation, the target logit is generated by feeding the current batch through the teacher network. This means the target logit for a sample varies based on the current batch statistics and data augmentations. Varying each sample's target logit across epochs significantly complicates analysis, so we aim to assign a single target logit per image. We find the best method is to average a sample's target logit across several batches and data augmentations.
\end{enumerate}

\subsection{Replica-based Mean Field Theory Manifold Analysis}
In this section, we summarize the theory underlying the Replica-based Mean-Field Theoretic Manifold Analysis (MFTMA) used in this paper. First introduced in \cite{chung2018classification}, this framework has been used to analyze internal representations of deep networks, ranging from visual \cite{cohen2020separability} to speech problems\cite{stephenson2019}. Based on replica theory in statistical physics, the MFTMA framework measures the "manifold" capacity, defined as the maximum number of object manifolds (point clouds in the feature space, tagged with categorical labels) such that the majority of the ensemble of random dichotomy labels for these objects manifolds can be linearly separated. This is a direct generalization of 'shattering' capacity of a perceptron, where the counting unit for the perceptron is number of objects (where each objects manifold can include a finite or infinite number of points), rather than number of discrete patterns. The expression for the manifold capacity in the MFTMA framework gives rise to new measures for characterizing geometric properties of object manifolds. Example of these object manifolds are population responses to different exemplars in the same object class (a class manifold), or an exemplar manifold created by a variability around a single stimulus (i.e, a single image or an utterance), where variability in the manifold can be from an adversarial perturbation, such that the shattering capacity of object manifolds can be formally expressed in terms of the geometric properties of object manifolds. As the measure of 'manifold capacity' can be empirically evaluated (just as the perceptron capacity can be empirically evaluated), the match between the empirical manifold capacity and the theoretical manifold capacity predicted from the object manifolds' properties has been shown in many domains with different datasets \cite{chung2018classification, cohen2020separability, stephenson2019}. As the framework formally connects the representational geometric properties and the object manifold's classification capacity, the measures from this framework are particularly useful for gaining a mechanistic account of how information content about objects are embedded in the structure of the internal representations from deep networks. Below we provide additional details of the measures from this framework: manifold capacity and the geometrical properties (such as manifold dimension, radius, width, and manifold center correlation).  

\subsubsection{Manifold Capacity}
Given neural or feature representations where $P$ object manifolds are embedded in $N$-dimensional ambient feature (or neural state) space, \textit{load} is defined as $P/N$. Large/small load implies that many/few object manifolds are embedded in the feature dimension. Consider a linear classification problem where binary positive and negative labels are assigned randomly to $P$ object manifolds, while all the points within the same manifold share the same label, and the problem is to find a linearly classifying hyperplane for these random manifold dichotomies. Note that there are $2^P$ manifold dichotomies, as each object manifold can be either assigned positive or negative labels. 

\paragraph{Manifold capacity} is defined as the critical load $\alpha_c = P / N$ such that above this value, most of these dichotomies have linearly separating solution, and below this value, most of the dichotomies do not have linearly separating solution. A system with a large manifold capacity has object manifolds that are well separated in the feature space, and a system with a small manifold capacity has object manifolds that are highly entangled in the feature space.  

\paragraph{Interpretation} Here we provide useful interpretations of manifold capacity. First, as the manifold capacity is defined as the critical load for a linear classification task, it captures the linear separability of object manifolds. Second, the manifold capacity is defined as the \textit{maximum} number of object manifolds that can be packed in the feature space such that they are linearly separable, it has a meaning of how many object manifolds can be "stored" in a given representation such that they can distinguished by the downstream linear readout. Third, the manifold capacity captures the amount of linearly decodable object information per feature dimension embedded in the distributed representation.  

Manifold capacity, $\alpha_{M} = P / N$, can be estimated using the replica mean field formalism with the framework introduced by \cite{chung2018classification} and refined in \cite{cohen2020separability}. As mentioned in the main text, $\alpha_{M}$, is estimated as $\alpha_{MFT}$, or MFTMA manifold capacity, from the statistics of \textit{anchor points}, $\tilde{s}$, a representative point for the points within a object manifold that contributes to a linear classification solution. The general form of the MFTMA manifold capacity has been shown \cite{chung2018classification, cohen2020separability} to be:

$$\alpha_{MFTMA}^{-1}=\left\langle \frac{\left[t_{0}+\vec{t}\cdot\tilde{s}(\vec{t})\right]_{+}^{2}}{1+\left\Vert \tilde{s}(\vec{t})\right\Vert ^{2}}\right\rangle _{\vec{t},t_0}$$

where $\left\langle \ldots\right\rangle _{\vec{t},t_0}$ is a mean over
random $D$- and 1- dimensional Gaussian vectors $\vec{t}, t_0$ whose components are i.i.d. normally distributed $t_{i}\sim\mathcal{N}(0,1)$. 

This framework introduces the notion of \textit{anchor points}, $\tilde{s}$, uniquely given by each $\vec{t}, t_0$, representing the variability introduced by all other object manifolds, in their arbitrary orientations. $\tilde{s}$ represents a weighted sum of support vectors contributing to the linearly separating hyperplane in KKT (Karush–Kuhn–Tucker) interpretation. 

\subsubsection{Manifold Geometric Measures}
The statistics of these anchor points play a key role in estimating a object manifold's effective Manifold Radius $R_M$ and Manifold Dimension $D_M$, as they are defined as: 

$$R_{\text{M}} =\sqrt{\left\langle \left\Vert \tilde{s}(\vec{T})\right\Vert ^{2}\right\rangle _{\vec{T}}}$$ 

$$D_{\text{M}}=\left\langle \left(\vec{t}\cdot\hat{s}(\vec{T})\right)^{2}\right\rangle _{\vec{T}}$$

 where $\hat{s} = \tilde{s} / \Vert \tilde{s} \Vert$ is a unit vector in the direction of $\tilde{s}$, and $\vec{T}=(\vec{t},t_0)$, representing a combined coordinate for the manifold's embedded space, $\vec{t}$, and the manifold's center direction $t_0$. 

\paragraph{Manifold Dimension} measures the dimensionality of the projection of $\vec{t}$ on its unique anchor point $\tilde{s}$, capturing the dimensionality of the regions of the manifolds playing the role of support vectors. In other words, the manifold dimension is the dimensionality of the object manifolds realized by the linearly separating hyperplane. High value of $D_M$ implies that the fraction of the part within the object manifold embedded in the margin hyperplane is high-dimensional, thereby implying that the classification problem is hard. 

\paragraph{Manifold Radius} measures the average norm of the anchor points,  $\tilde{s}(\vec{T})$, within the manifold subspace, capturing the size of the object manifold realized by the linearly separating hyperplane. A small value of $R_M$ implies tightly grouped anchor points. 

\paragraph{Center Correlation} $\rho_{center}$ measures the average pairwise correlation between manifold centroids. Small correlation indicates that on average, manifolds lie along different directions in feature space.

If the object manifold centers are in random locations and orientations, the geometric properties predict the MFTMA manifold capacity \cite{chung2018classification}, by

$$\alpha_{\text{MFTMA}}\approx\alpha_{\text{Ball}}\left(R_{\text{M}},\,D_{\text{M}}\right)\approx\alpha_{\text{point}}\left(R_{\text{M}}\cdot \sqrt{D_{\text{M}}}\right)$$
where, 

$$\alpha_{\text{Ball}}^{-1}(R,D)=\int_{-\infty}^{R\sqrt{D}}Dt_{0}\frac{(R\sqrt{D}-t_{0})^{2}}{R^{2}+1}$$ 
is a capacity of $L_2$ spheres with radius $R$ and dimension $D$ as defined in \cite{chung2016linear} and 
$$\alpha_{\text{Point}}^{-1}(\kappa)=\int_{-\infty}^{\kappa}Dt(t-\kappa)^2 $$ 

is a classification capacity of points given an imposed margin of $\kappa$ as defined in \cite{chung2018classification}.

In real data, the manifolds have various correlations, hence the above formalism has been applied to the data projected into the null spaces of manifold centers, similar to the method proposed by \cite{cohen2020separability}. 

The validity of this method is shown in various literature \cite{cohen2020separability, stephenson2019}, where a good match between the MFTMA manifold capacity and the ground truth (empirical) capacity has been demonstrated (where the empirical capacity is computed using a bisection search on the critical number of feature dimensions required in order to reach roughly ~50\% chance of having linearly separable solutions given a fixed number of manifolds and their geometries). Hence, the manifold capacity or $\alpha_M$ refers to the mean-field approximation of the manifold capacity, $\alpha_{MFTMA}$ or MFTMA Manifold Capacity. For more details on the theoretical derivations and interpretations for the mean-field theoretic algorithm, see \cite{cohen2020separability, chung2018classification}.

\subsubsection{Capacity: theory vs. simulation}
\begin{figure}[h!]
  \centering
    \begin{tikzpicture}
    \pgftext{\includegraphics[width=0.8\columnwidth]{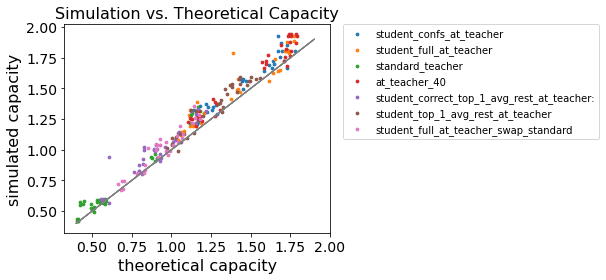}};
    \end{tikzpicture}
          \vspace{-4mm}
  \caption{\textbf{MFTMA capacity approximately predicts simulated capacity} for all layers and all models presented, consistent with prior work \cite{cohen2020separability}.}
  \label{fig:simcap_vs_mftma}
\end{figure}

In Figure \ref{fig:simcap_vs_mftma} we show empirical (simulated) capacity vs MFTMA predicted capacity for adversarial perturbation manifolds in all analyzed layers of the models presented in the current study. We observe a tight relationship between MFTMA and simulated capacity, with MFTMA mildly overestimating capacity consistent with prior work \cite{cohen2020separability}.

\section{Additional results}
\label{sec:additional}
\setcounter{figure}{0}
\subsection{Teacher logits}

\begin{figure}[h]
    \centering
    \includegraphics{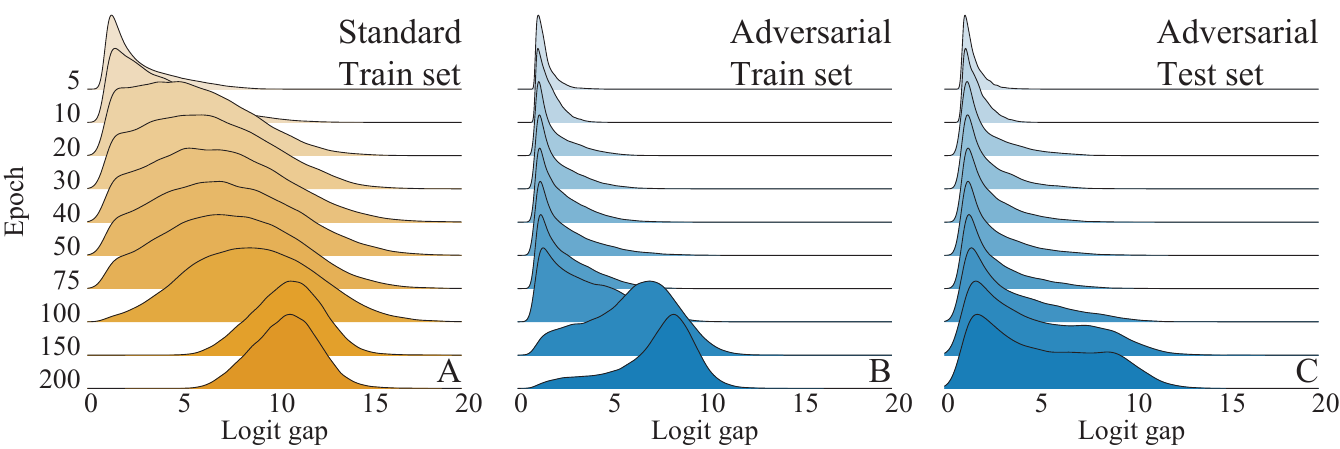}
    \caption{Logit gap distributions across training epochs. (A, B) ST and AT logit gap distributions on the training set become very similar after epoch 100. (C) AT logit gap distribution on the testing set maintains a highly skewed, lower mean distribution.}
    \label{fig:ridgeline-gap}
\end{figure}

We find that an AT network can no longer transfer robustness after epoch 100 and this is likely due to the maximum logit and logit gap distributions becoming similar to the ST network at this late epochs. Figure \ref{fig:ridgeline-gap} A and B illustrate that the ST and AT logit distributions are similar on the training set at late epochs. But these late epoch AT networks are still robust, so there should be some distribution differences to ST network. Figure \ref{fig:ridgeline-gap} C shows that the AT network maintains a significantly skewed distribution on the testing set after epoch 100, indicating the AT network is still robust. 

\begin{figure}[H]
    \centering
    \includegraphics{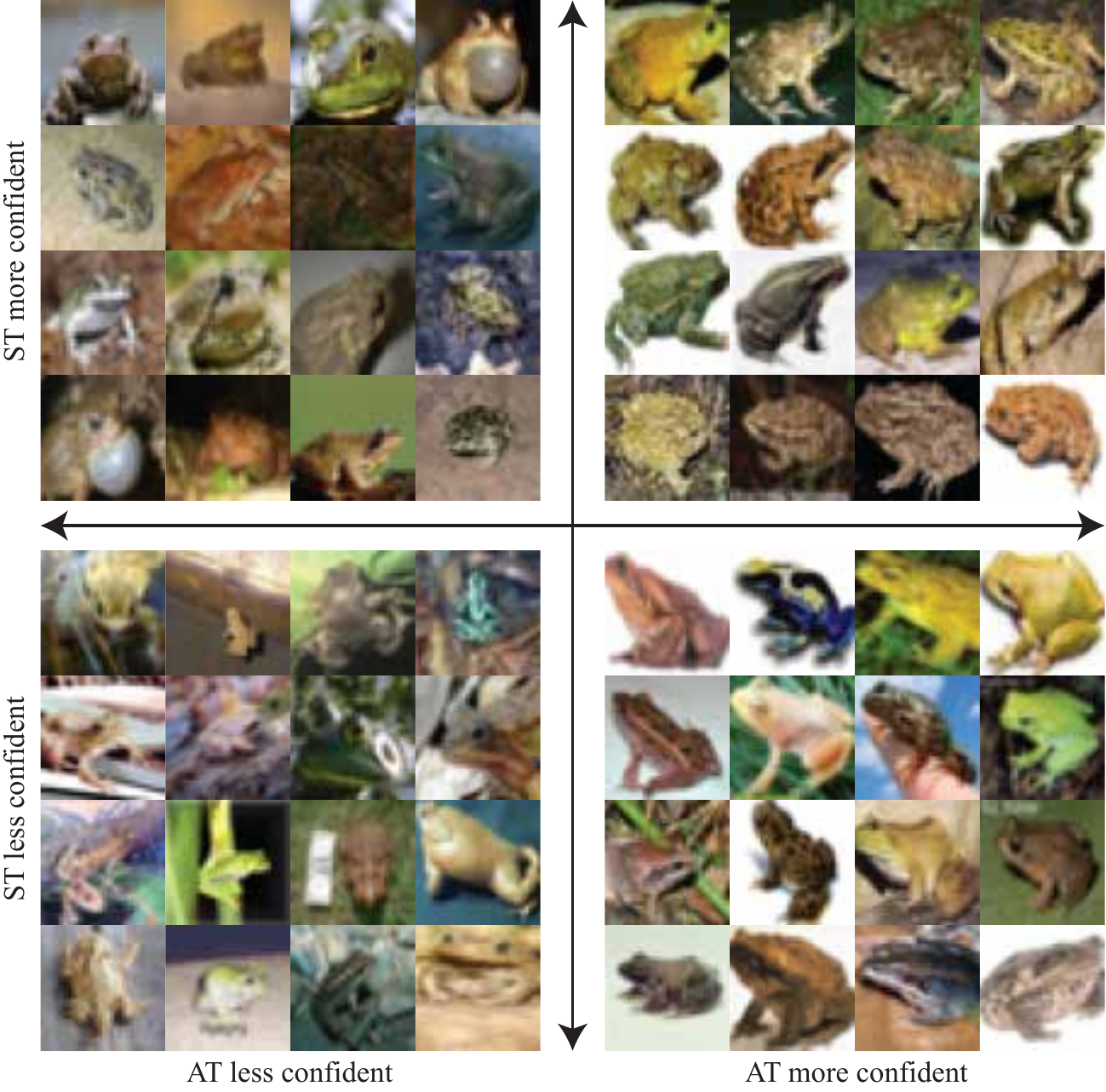}
    \caption{ST and AT networks are confident in their responses to different frog images. AT networks are primarily confident in unambiguous samples where the object's shape is prominent (right quandrants). ST networks are confident in a more diverse set of samples (upper quadrants), with some plain background images (top right) but many more complex frog images (top left).}
    \label{fig:confidence_images}
\end{figure}

We demonstrated that AT and ST networks are confident in different types of images in Figure \ref{fig:logit_specifics}B. In Figure \ref{fig:confidence_images}, we visualize samples where the AT and ST networks either agree or disagree on high and low confidence images. From the top-right and bottom-left squares, AT and ST networks agree on white background images being high confidence and ambiguous images being low confidence. From the bottom-right and top-left squares, the ST network appears to label some white background images as lower confidence and prioritizes more ambiguous samples to be high confidence. This shows AT networks are more consistent at rating samples with clear class indicators as high confidence.

\begin{figure}[h]
    \centering
    \includegraphics[width=\textwidth]{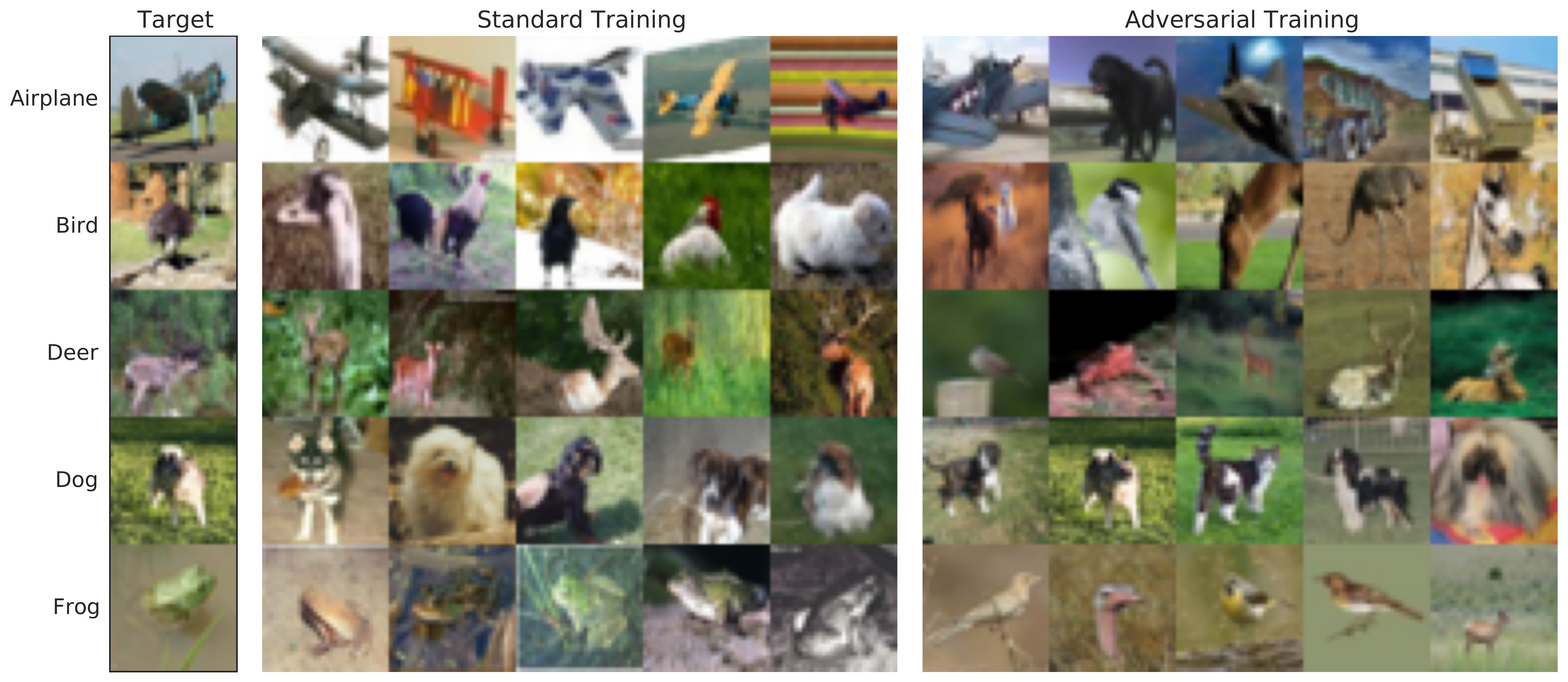}
    \caption{AT networks represent visually similar images more closely together than ST networks. The left column shows seed images from five CIFAR10 categories. Each row shows the 5 most similar images according to an ST network (center) and AT network (right) to the corresponding seed image, as ranked by the cosine similarity of the logits. AT networks favor similar colors and shapes, at the expense of keeping images from different classes close together.}
    \label{fig:similarity}
\end{figure}

\subsection{CIFAR100}
\label{sec:cifar100}

\begin{figure}[h]
\begin{subfigure}{.5\textwidth}
    \centering
    \includegraphics[width=\textwidth]{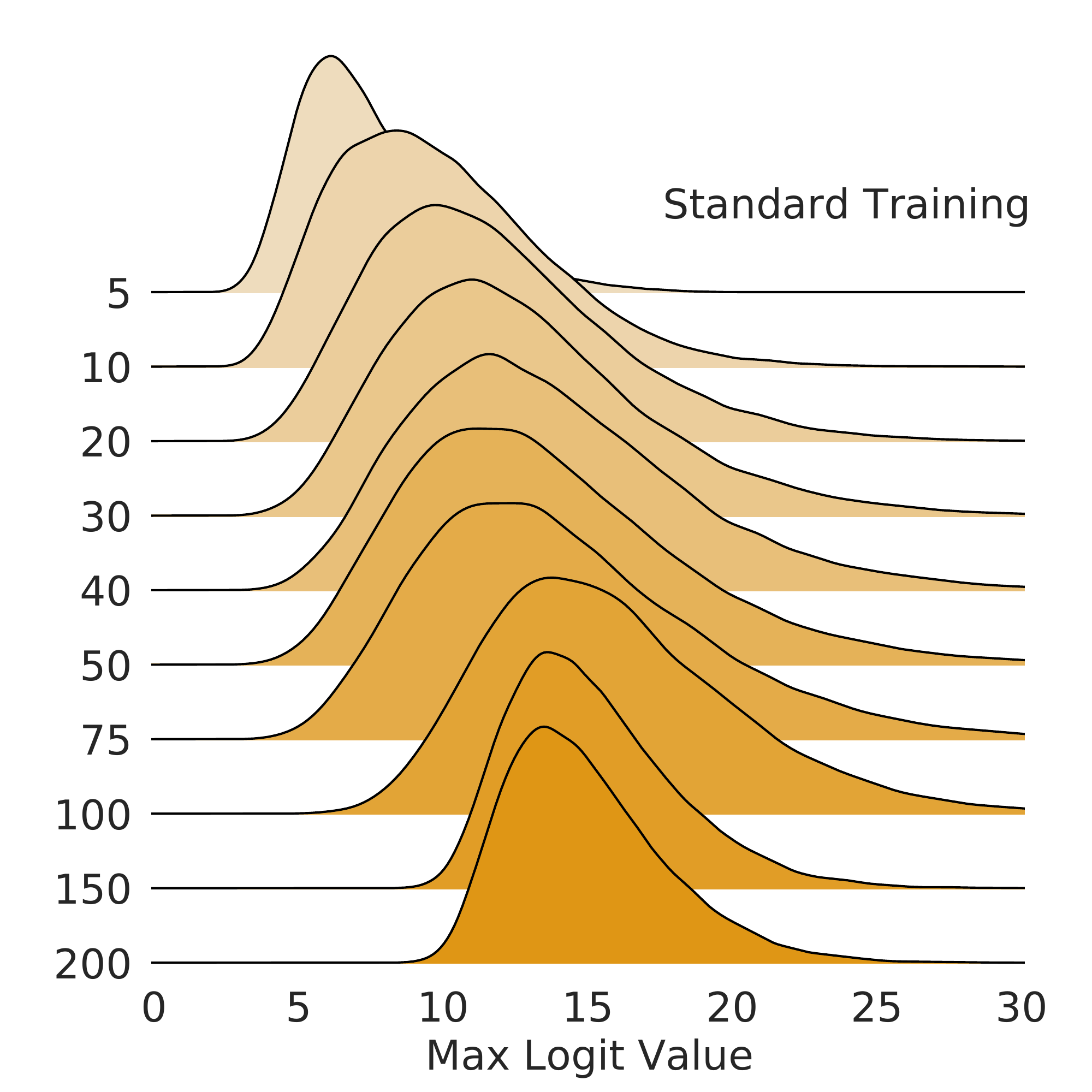}
    \caption{}
    \label{fig:cifar100_ST_max_logit}
\end{subfigure}
\begin{subfigure}{.5\textwidth}
    \centering
    \includegraphics[width=\textwidth]{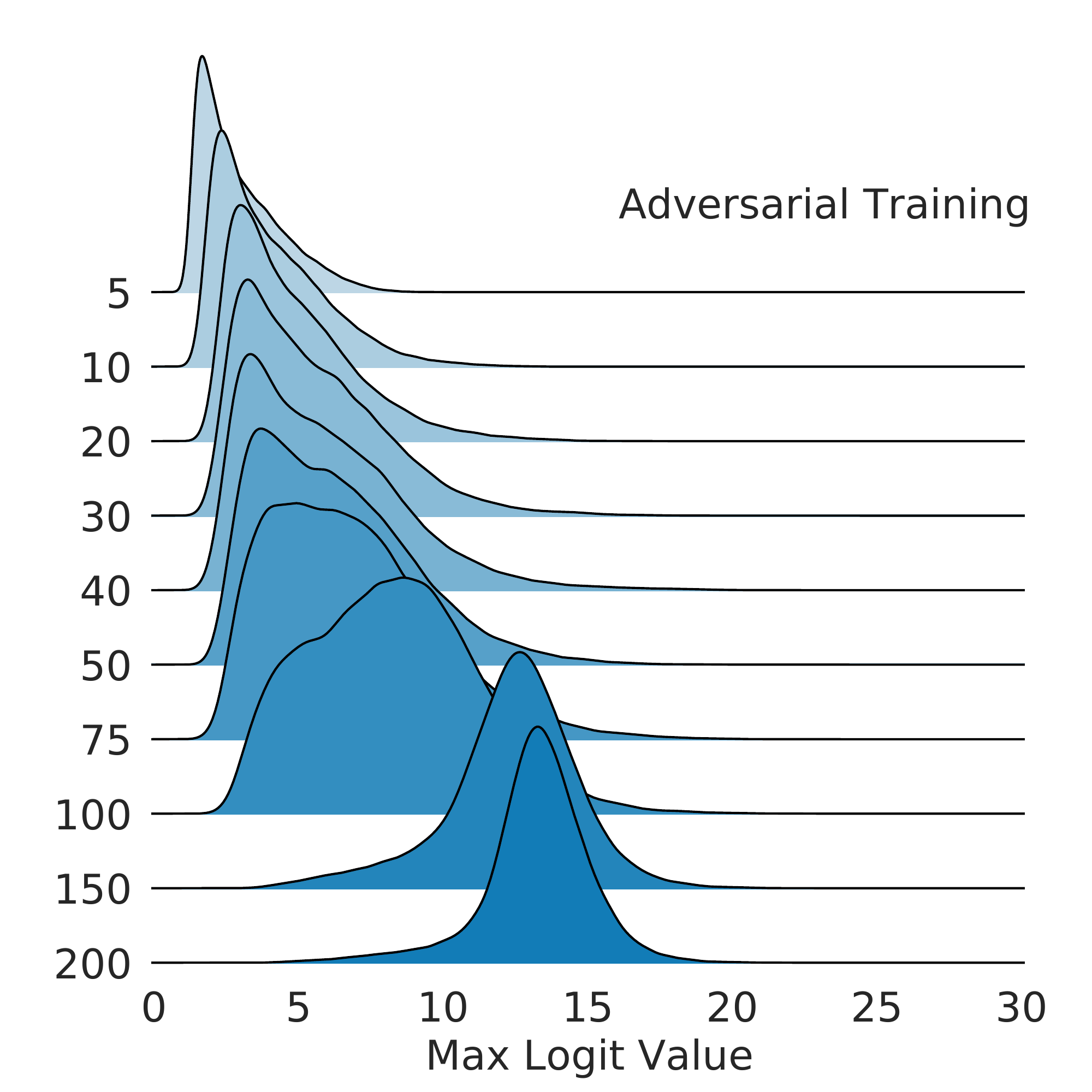}
    \caption{}
    \label{fig:cifar100_AT_max_logit}
\end{subfigure}
\caption{Evolution of the maximum logit values through 200 epochs of training for CIFAR-100. (a) Standard training results in distributions with less skew and a significantly larger mean. (b) Adversarial training results in distributions with significantly more positive skew and lower mean until very late in training.}
\label{fig:cifar100_max_logit}
\end{figure}

Here, we replicate the maximum logit value analysis for CIFAR-100. Figure \ref{fig:cifar100_max_logit} compares how standard training (ST) and adversarial training (AT) effect a network's max logit value for the CIFAR-100 dataset. Similar to Figure \ref{fig:logit_max}, AT networks are more positively skewed with a long positive tail early in training. Eventually, AT networks converge to an almost symmetric distribution. In contrast to the CIFAR-10 results, the ST network is less symmetric and exhibits a slightly longer positive tail. Additionally, both ST and AT models have larger average maximum logits.


\begin{figure}[h!]
\begin{subfigure}{.5\textwidth}
    \centering
    \includegraphics[width=\textwidth]{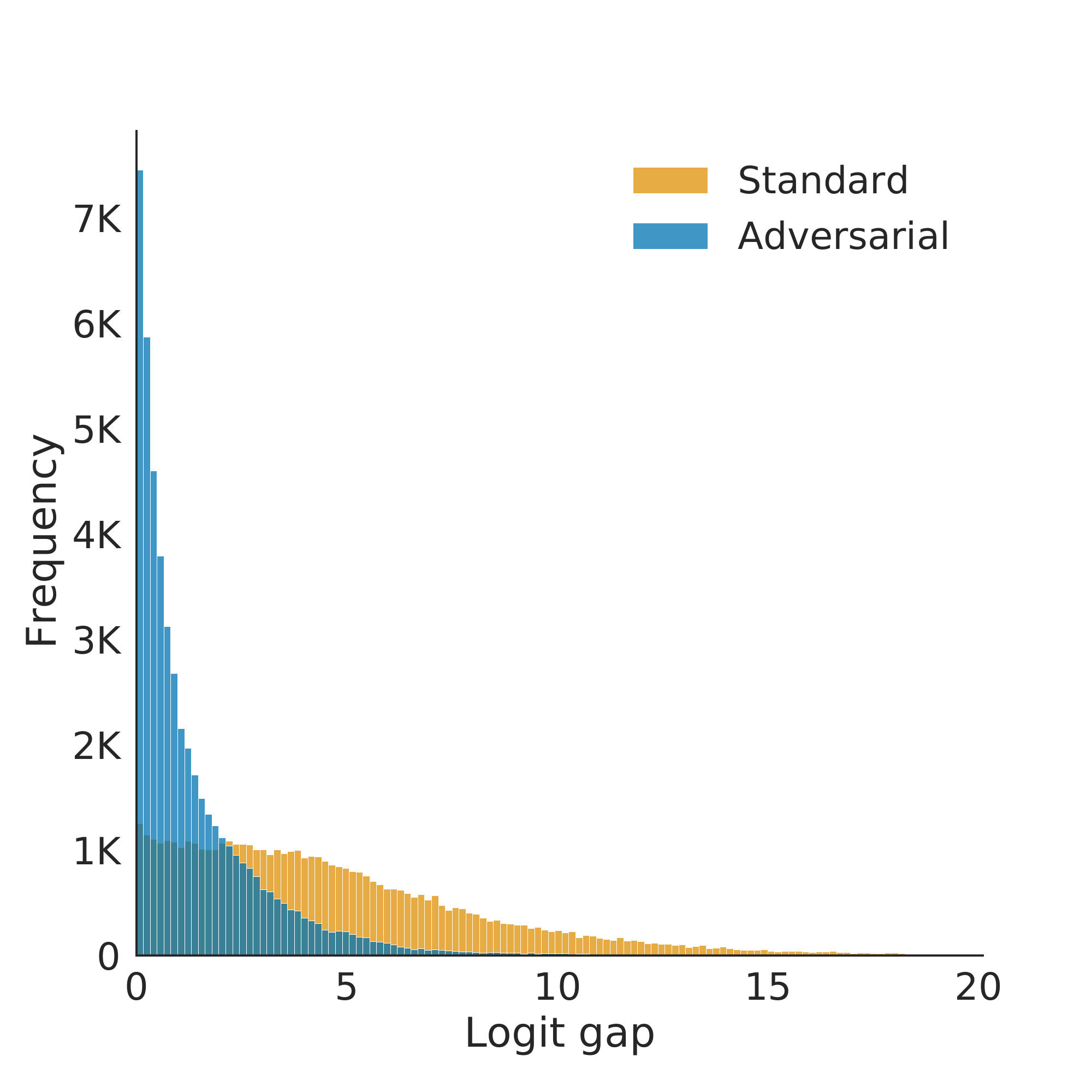}
    \caption{}
    \label{fig:cifar100_compar_gap}
\end{subfigure}
\begin{subfigure}{.5\textwidth}
    \centering
    \includegraphics[width=\textwidth]{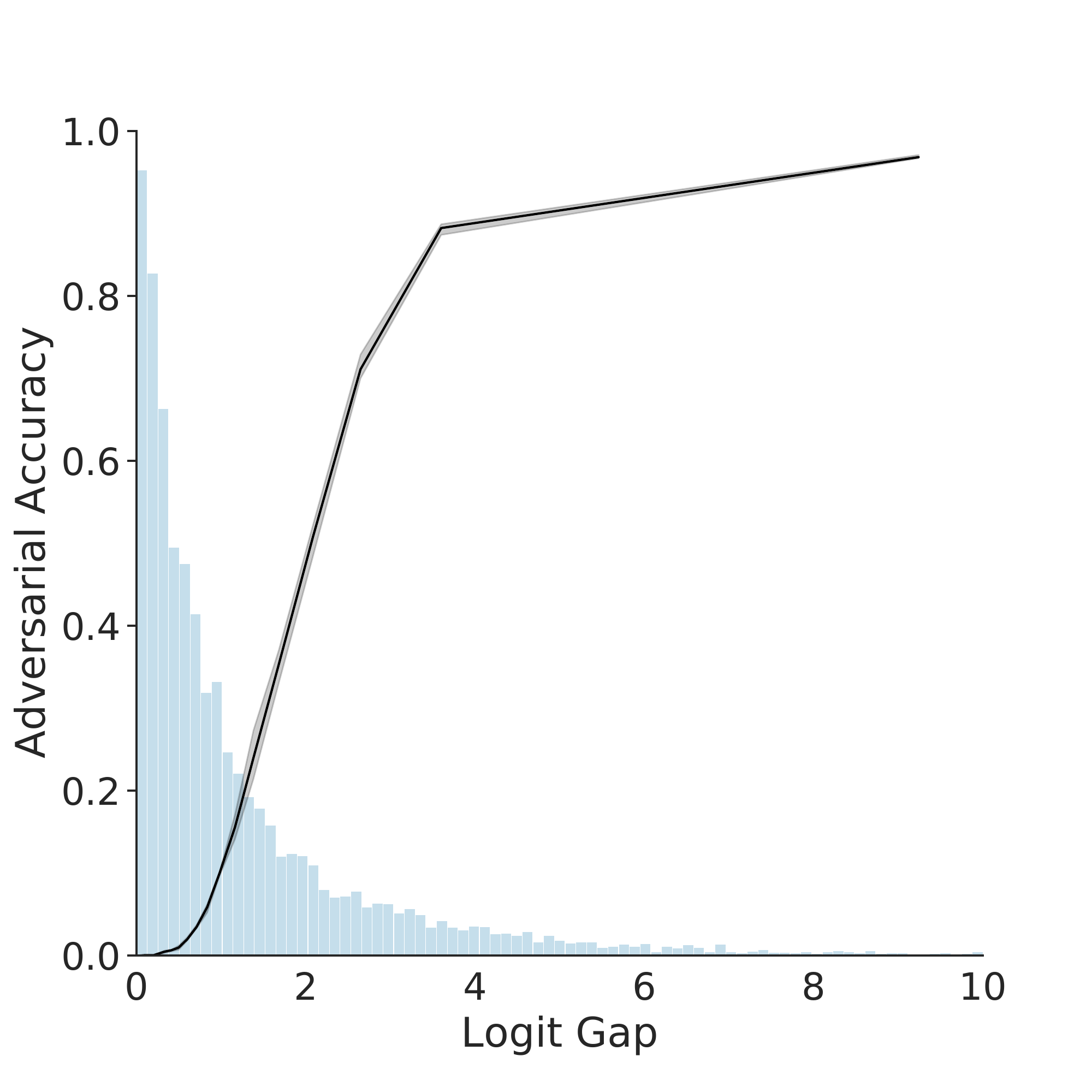}
    \caption{}
    \label{fig:cifar100_acc_v_gap}
\end{subfigure}
\caption{Comparing the logit gaps of adversarial trained (AT) networks and standard trained (ST) networks at epoch 40. (a) AT network logit gaps are significantly smaller than ST networks with a peak at zero and few gaps larger than five. (b) The black line with shaded region represents the adversarial accuracy mean and standard deviation for three networks across logit gaps. In contrast to ST networks (with zero adversarial accuracy), AT network's large logit gaps are a strong indicator of adversarial accuracy.}
\label{fig:cifar100_gap}
\end{figure}

Next, we examine logit gap distribution for CIFAR-100. Figure \ref{fig:cifar100_gap} compares the logit gaps of AT and ST models at epoch 40 of training. Similar to the CIFAR-10 results in Figure \ref{fig:logit_gap}, the CIFAR-100 AT network has a logit gap distribution with a distinct peak at zero and a quick decline for larger values. The CIFAR-100 ST network still has significantly larger logit gaps than the AT network, but the mean logit gap is noticeably less than the mean logit gap for CIFAR-10 ST networks. This shift is likely due to the increased difficulty of CIAR-100.

\begin{figure}[h!]
    \centering
    \includegraphics[width=\textwidth]{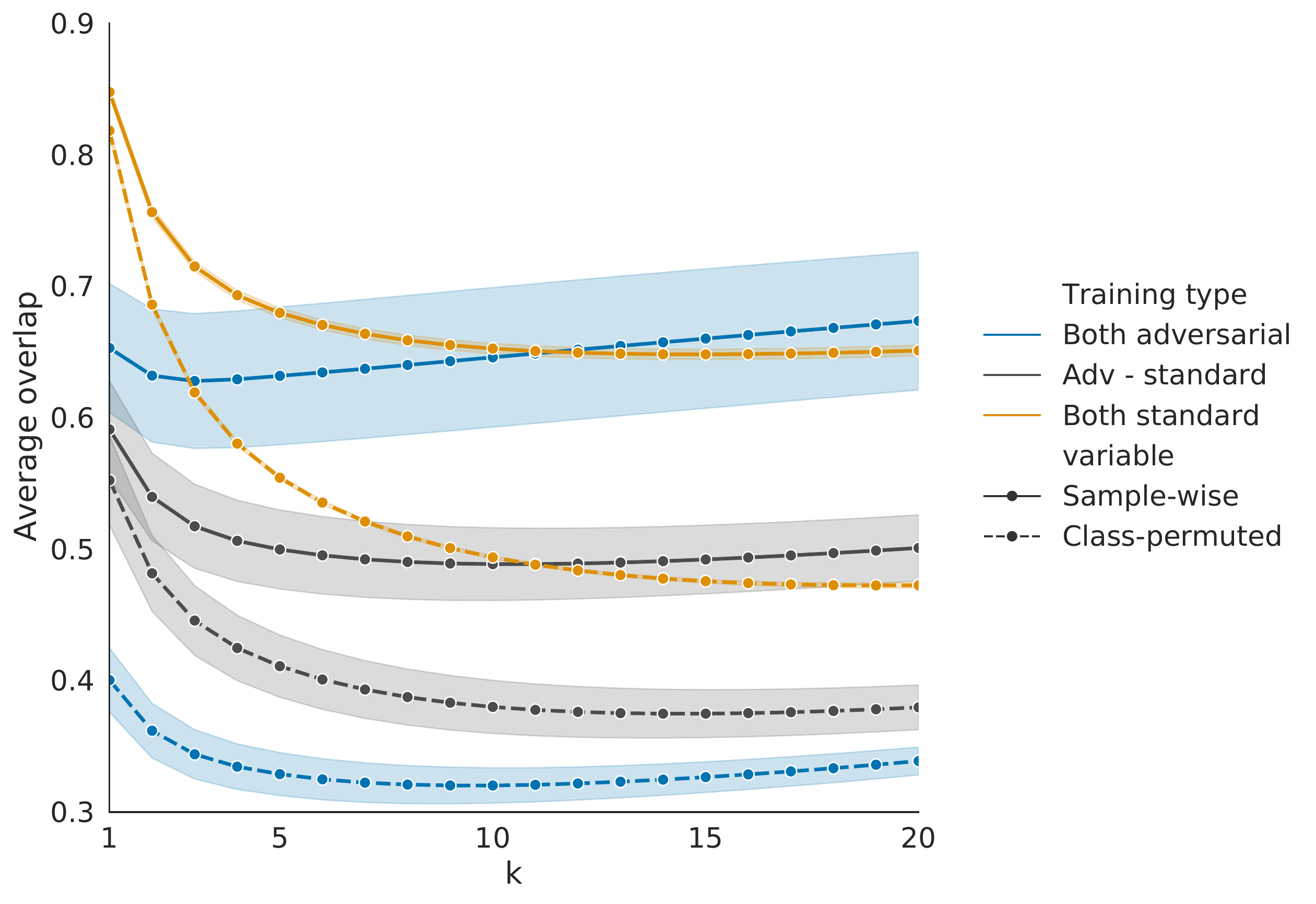}
    \caption{Measuring the "average overlap" statistic to compare the set of classes predicted on individual examples by ST and AT networks (solid lines). AT networks have higher overlap with other AT networks than with ST networks, suggesting that AT networks select different sets of likely classes than ST networks. This difference is reduced if examples are compared with other images from the same class (dashed lines), suggesting that AT networks provide more diverse predictions to images from the same class. All comparisons are plotted as mean $\pm$ standard deviation.}
    \label{fig:cifar100-avg-overlap}
\end{figure}

Finally, we computed the average overlap \cite{webber2010similarity} between the logit rank orders for AT and ST networks on CIFAR-100. In contrast to CIFAR-10, we find ST networks to have more overlap than AT networks for $k < 10$. After $k > 10$, AT networks start to have more overlap than ST networks. The AT network overlap has a larger degradation than ST network overlap when we permute the samples within each class. Similar to CIFAR-10, this illustrates AT networks have more output variability in response to examples of the same class. The results are shown in Figure \ref{fig:cifar100-avg-overlap}.

\clearpage
\section{An Analytically Tractable Model}
\setcounter{figure}{0}
\label{sec:analytical}

\newcommand{\dg}{^{T}}
\newcommand{\R}{\mat{R}}
\newcommand{\vst}{\par\vspace{2.5mm}}
\newcommand{\be}{\begin{equation}}
\newcommand{\ee}{\end{equation}}
\newcommand{\bea}{\begin{eqnarray}}
\newcommand{\eea}{\end{eqnarray}}
\newcommand{\beNN}{\begin{equation*}}
\newcommand{\eeNN}{\end{equation*}}
\newcommand{\beaNN}{\begin{eqnarray*}}
\newcommand{\eeaNN}{\end{eqnarray*}}
\newcommand{\Oh}{\mbox{\large$\mathpzc{O}$}}

\newcommand{\Xx}{\mat{X}}
\newcommand{\Yy}{\mat{Y}}
\newcommand{\micro}{\fontsize{2.25}{2.25}\selectfont}
\newcommand{\mini}{\fontsize{4.5}{4.5}\selectfont}
\newcommand{\smaller}{\fontsize{7}{7}\selectfont}
\newcommand{\smallerA}{\fontsize{7.5}{7.5}\selectfont}
\newcommand{\smallerB}{\fontsize{8.5}{8.5}\selectfont}
\newcommand{\smallerC}{\fontsize{9.5}{9.5}\selectfont}
\newcommand{\medium}{\fontsize{11}{11}\selectfont}
\newcommand{\gargantuan}{\fontsize{30}{30}\selectfont}
\newcommand{\Gargantuan}{\fontsize{40}{40}\selectfont}
\newcommand{\smb}{\smallerB}
\newcommand{\Ndata}{{N_{\mathrm{\smaller{data}}}}}
\newcommand{\Nclasses}{{N_{\mathrm{\smaller{classes}}}}}
\newcommand{\Nfeats}{{N_{\mathrm{\smaller{feats}}}}}
\newcommand{\scr}{\mathscr}
\newcommand{\mat}{\mathbb}

\newcommand{\lmsd}{\mathbf}

\newcommand{\tabL}{\begin{enumerate}}
\newcommand{\tabR}{\end{enumerate}}
\def\vekh#1{\vek{\hat{#1}}}

\newcommand{\Zee}{\mathpzc{Z}}
\newcommand{\lnZ}{\log{\!\Zee}}
\def\CE#1#2{\mathscr{L}_{CE}{ (#1}\vert{#2 )}}
\def\vek#1{{\boldsymbol{#1}}}
\newcommand{\dejavu}{\mathrm}
\newcommand{\Diag}{\dejavu{DIAG}}

\newcommand{\Tr}{\mathrm{Trace}}
\newcommand{\smallperp}{\mbox{\micro{$\perp$}}}
\newcommand{\maxdir}{\vekh{\phi}_\lmsd{\tiny z}}
\def\slotX#1{{\vek{X}^{[#1]}}}
\newcommand{\logitvec}{\lmsd{z}}
\newcommand{\logitvect}{\tilde{\lmsd{z}}}
\newcommand{\logitvecmu}{\lmsd{logit}^{[\mu]}}
\newcommand{\logitperp}{{^{\smallperp}\!\!\logitvec}}
\newcommand{\logitPerp}{{^{\smallperp}\logitvec}}
\newcommand{\logitperpmu}{^{\perp}\logitvecmu}
\newcommand{\logitmax}{\widehat{\lmsd{max}}}
\newcommand{\logitmaxmu}{\widehat{\lmsd{max}}^{[\mu]}}
\newcommand{\vperp}{{^{\smallperp}\lmsd{v}}}

\newcommand{\betaParallel}{\beta_{\parallel}}
\newcommand{\betaStar}{{\beta_\star}^{\!\!\!\parallel}}
\newcommand{\betaParallelmu}{\betamu^{\!\!\!\parallel}}
\newcommand{\fPerp}{\dejavu{f}}
\newcommand{\gPerp}{\dejavu{g}}
\newcommand{\fperpPlus}{\fPerp_{+}}
\newcommand{\fperpMinus}{\fPerp_{-}}
\newcommand{\ones}{\lmsd{ones}}
\newcommand{\veky}{\lmsd{y}}
\newcommand{\vekx}{\lmsd{x}}
\newcommand{\betaz}{{\beta_{\smaller\logitvec}}}
\newcommand{\betazmu}{{\betaz}_{\mu}}
\newcommand{\hbetaz}{\vek{h}_{\betaz}}
\newcommand{\wrong}{{{\mbox{\smaller\ding{55}}}}}
\newcommand{\correct}{{\checkmark}}
\newcommand{\betaWrong}{{\beta_\wrong}}
\newcommand{\betaCorrect}{{\beta_\correct}}
\newcommand{\error}{\varepsilon_{\mathrm{error}}}
\newcommand{\Jac}{\dejavu{Jac}}
\def\slotJac#1{\left[\Jac_{\smallerB\vekx^{#1}}\logitvec^{#1}\right]}
\newcommand{\ess}{\scr{S}}
\newcommand{\Zz}{\mat{Z}}
\newcommand{\Zzt}{\mat{\tilde{Z}}}


\begin{table}[H]
\label{sample-table}
\begin{center}
\begin{tabular}{lllllll}
 & description & dimensions
\\ \hline \\
$\veky$     & class index represented as a one-hot vector  &  $\Nclasses$  \\
$\logitvec(\vekx)$  & logit given input sample $\vekx$ & $\Nclasses$   \\
$\betaz$  & $\logitvec\cdot\max{()}$, i.e. the largest component of $\logitvec$ & 1   \\
$\CE{\logitvec}{\veky}$ & cross-entropy loss given $(\logitvec, \veky)$  & 1\\
$\vek{a}\odot\vek{b}$ & elementwise product of the vectors $\vek{a},\,\vek{b}$  & dimension of $\vek{a}$\\
$\vek{a}\otimes\vek{b}$ & outer product of the vectors $\vek{a},\,\vek{b}$  & dimension of $\vek{a}
$ $\times$ dimension of $\vek{b}$\\
$\Diag(\vek{a})$ & diagonal matrix from the elements of $\vek{a}$  &(dimension of $\vek{a}$)$^2$\\
$\vek{1}$ & vector whose components are all unity &\\
\end{tabular}
\end{center}
\end{table}

\subsection{Preliminaries}
\begin{itemize}
\item
Given any vector $\vek{v}\in\R^m$,  $m\ge2$, define the subset $\ess_{m}(\vek{v})\subset\R^m$ via 
\be
\ess_m(\vek{v}):=
\left\{\vek{u}\in\R^m \,\Big\vert\, \vek{v}\cdot\vek{u}=0,
\,\,\max_{1\le i\le m}{\{u_i\}} < \max_{1\le i\le m}{\{v_i\}}
\right\}
\label{Sset}
\ee

In other words, $\ess_{m}(\vek{v})$ consists of all vectors in $\R^m$ which are orthogonal 
to $\vek{v}$ and have their largest components strictly smaller than the largest component of 
$\vek{v}$.

\item[]

\item
Given any {\em trained} deep neural network with logits $\logitvec(\vekx)$ corresponding to training input 
$\vekx$ with label $\veky(\vekx)$, we define 
\begin{enumerate}
\item[-]
$ \betaz(\vekx):=\mbox{largest component of $\logitvec(\vekx)$}$
\item[-]
$\maxdir(\vekx):=$ projection of  $\logitvec(\vekx)$ onto its largest component, i.e. 
$\maxdir(\vekx)\cdot\logitvec(\vekx) = \betaz(\vekx)$.
\end{enumerate}
\end{itemize}

\subsection{Analytically tractable logit estimates}
\subsubsection{Motivation}
A deep neural network is trained to construct a mapping $\vekx\mapsto\logitvec(\vekx)$ via a complex 
composition of parametrized non-linear functions, rendering any analytical treatment intractable. Our approach proposes replacing the actual trained logits $\logitvec(\vekx)$ with surrogates that, in a very precise sense, mimic the true logits. 

Given a training sample $\big(\vekx, \veky(\vekx)\big)$ that yields a logit $\logitvec(\vekx)$, we fix the  projection $\betaz\maxdir$ onto the logit's largest component, and consider the set $\ess_{\Nclasses}(\maxdir\betaz)$ (c.f. defintion\,(\ref{Sset})). Within each such set, we minimize the cross-entropy loss and extract

\[
\vperp(\betaz(\vekx)\maxdir(\vekx), \veky(\vekx)):=
\argmin
_{\vek{u}\in\ess_{N_c}\!(\maxdir\betaz)}
\CE{\betaz(\vekx)\maxdir(\vekx)+\vek{u}}{\veky(\vekx)}
\]

Once this is done, we will have a parametrization of surrogate logits $\logitvect$ such that  
\tabL
\item[-]
$\tilde{\logitvec}(\vekx) :=\betaz(\vekx)\maxdir(\vekx)+\vperp(\betaz(\vekx),\maxdir(\vekx),\veky)$ is a stationary point 
for the loss, and 
\item[-]
the largest component of $\tilde{\logitvec}(\vekx)$ and the original/true $\logitvec(\vekx)$ are identical.
\tabR

It follows that the surrogate logits $\logitvect$ and the true logits  $\logitvec$ {\em have identical
(top-1) classification performance on the training set, and both lie in neighborhoods that contain stationary/critical points of the cross-entropy loss}.

Thus, given logits $\logitvec$ from a trained deep neural network, e.g. logits from a Resnet trained on CIFAR-10, we carry out the minimization procedure outlined above to obtain a set of surrogate logits  $\logitvect$, such that the latter provide us with analytically tractable estimates of 
former. 

\subsubsection{Constructing surrogate logits}
Our ultimate goal is to derive an analytically tractable model for the logits of a trained deep neural net {\em under the assumption that the neural net achieves an error rate $\varepsilon_{\mathrm{error}}$ in the absence of adversarial perturbations to the inputs}.

Following the discussion in the previous section, for a trained model with an error rate $\error$, we have, by definition, that for input $\vekx$ with label $\veky(\vekx)$

\tabL
\item[-]
$
\maxdir(\vekx) \parallel \veky(\vekx) \mbox{ with frequency $1-\error$}
$
\item[-]
$
\maxdir(\vekx) \perp \veky(\vekx) \mbox{ with frequency $\error$}
$
\tabR

Crucially, we note that the error rate $\error$ is not an independent free parameter, but rather a function of $\logitvec$ and $\veky$. 

\subsubsection{Loss minimization}
As previously discussed, we construct candidates for surrogate logits by 
solving the minimization problem 

\be
\vperp(\betaz(\vekx)\maxdir(\vekx), \veky(\vekx)):=
\argmin_{\vek{u}\in\ess_{N_c}\!(\maxdir\betaz)}
\CE{\betaz(\vekx)\maxdir(\vekx)+\vek{u}}{\veky(\vekx)}
\ee

We carry out the minimization procedure separately for each of the following cases:

\begin{itemize}
\item
If $\logitvec(\vekx)$ leads to the correct classification of the input training pair $(\vekx, \veky(\vekx))$, we are assured that  $\maxdir(\vekx) \parallel \veky(\vekx) $, in which case, we minimize 

\[
\CE{\betaz\veky+\vek{u}}{\veky}
\,\,
 \mbox{ over all \,\,$\vek{u}\in\ess_{\Nclasses}(\betaz\veky)$}
\]

\item
If, instead, the input training pair $(\vekx, \veky)$ is misclassified, we have  $\maxdir(\vekx) \perp \veky $. For such instances, we minimize 

\[
\CE{\betaz\maxdir+\vek{u}}{\veky}
\,\,
\mbox{ over all $ \,\,\vek{u}\in\ess_{\Nclasses}(\betaz\maxdir)$ }
\]

\end{itemize}

\subsubsection{Perturbative solutions to the minimization problem}
For analytical tractability, we assume that the logits of a trained deep neural net are such that 
$\lVert \logitvec - \betaz\maxdir\rVert^4 \ll \vert\betaz\vert^4$. Intuitively, this means that the largest component of $\logitvec$ and its norm are roughly of the same size. This allows us to analyze the loss as a convergent power series of  $\lVert \logitvec - \betaz\maxdir\rVert$. A straightforward Taylor expansion \footnote{There are several reasons why this approach is expected to yield insight into the complete optimization of the loss, one of which is outlined below.} of the loss in a neighborhood of $\betaz\maxdir$ yields

\bea
\CE{\betaz\maxdir+\vperp}{\veky}
&=&
\lnZ_{\betaz} 
-\betaz\maxdir\cdot\veky
+
\vperp\cdot(\hbetaz-\veky)
+
\frac{1}{2}
\vperp\cdot Q(\hbetaz)\vperp
\nonumber \\
&&
\qquad
+
\frac{1}{6}
(\vperp\odot\vperp)\cdot Q(\hbetaz)\vperp
-\frac{1}{3}
\left(
\vperp\cdot\hbetaz
\right)
\vperp\cdot Q(\hbetaz)\vperp
\nonumber\\
&&
\qquad\qquad
+
\Oh(\lVert\vperp\rVert^4)
\label{loss}
\eea

where 
$\lnZ_{\betaz} 
:=
\betaz + 
\ln{\left[
{1+e^{-\betaz}(\Nclasses-1)}
\right]}
$, and the external field $\hbetaz$, is given by 
\[
\hbetaz
:=
\frac{1}{e^\betaz+(\Nclasses-1)}
\left[
\vek{1}+(e^\betaz-1)\maxdir
\right]
\]

while $Q$ is a quadratic form defined as 
\[
Q({\hbetaz}):=
\Diag{(\hbetaz)}-\hbetaz\otimes\hbetaz
\]

After carefully walking through a lengthy, yet straightforward, calculation, one can fully  
characterize the local minima of the loss in equation (\ref{loss}). 

\begin{enumerate}
\item[(A)]
For correctly classified examples, the stationary points of (\ref{loss}) are given by

\bea
\vperp_{\pm}(\betaz, \veky)
&=&
\fPerp_{\pm}(\betaz)(\vek{1} -\veky) 
\,\,\,\mbox{with}\,\,\,
\betaz \in \{u\in\R\vert u>  \fPerp_{\pm}(u) \}
\label{logitsCase1}
\eea

where 

\beaNN
\fPerp_{\pm}(\betaz)
&:=&
\frac{1+e^{-\betaz}(\Nclasses-1)}{1-e^{-\betaz}(\Nclasses-1)}
\left[
-1\pm
\sqrt{
1+
\frac{2e^{-2\betaz}[1-e^{-\betaz}(\Nclasses-1)]}
{\big[1+e^{-\betaz}(\Nclasses-1)\big]^2
}}\right]
\eeaNN

\item[(B)]
For misclassified samples, i.e. $\maxdir \ne \veky$, the minima are given by

\bea
\vperp_{\pm}(\betaz, \maxdir, \veky)
&=&
\gPerp_{\pm}(\betaz)\veky
+
\psi_{\pm}(\betaz)(\vek{1} -\maxdir-\veky)
\,\,\,\mbox{with}\,\,\,
\betaz \in \{u\in\R\vert u > 
 \max{\left( \psi_{\pm}(u),  \gPerp_{\pm}(u) \right) }
\}
\nonumber \\
\label{logitsCase2}
\eea

where

\beaNN
\gPerp_{\pm}(\betaz)
&:=&
-
\frac{1-e^{-\betaz}(\Nclasses-3)}{1-e^{-\betaz}(\Nclasses-1)}
\big[
1\pm\kappa(\betaz)
\big]
\\
\kappa(\betaz)
&:=&
\frac{1+e^{-\betaz}(\Nclasses-1)}{1-e^{-\betaz}(\Nclasses-3)}
\sqrt{
1+
\frac{2 
[1-e^{-\betaz}(\Nclasses-1)][1-e^{-\betaz}(\Nclasses-3)]
}
{\big[1+e^{-\betaz}(\Nclasses-1)\big]
}
}
\\
\psi_{\pm}(\betaz)
&:=&
2\left[
\frac{e^{-\betaz}}{1-e^{-\betaz}(\Nclasses-1)}
\gPerp_{\pm}(\betaz)
-
\frac{1+e^{-\betaz}(\Nclasses-1)}{1-e^{-\betaz}(\Nclasses-3)}
\right]
\eeaNN

\end{enumerate}

The solutions in (\ref{logitsCase1},\,\ref{logitsCase2}) indicate that not all values of 
$\betaz$ give rise to stationary points of the loss. Rather,  candidates in 
$\ess_\Nclasses(\betaz\maxdir)$ only yield stationary points for the loss  
at select values of $\betaz$ in the intervals indicated in equations (\ref{logitsCase1},\,\ref{logitsCase2}). We can readily visualize permissible values of $\betaz$ by plotting 
the $\betaz$-constraints in equations (\ref{logitsCase1},\,\ref{logitsCase2}).

\begin{figure}[H]
\centering
\includegraphics[width=0.8\linewidth]{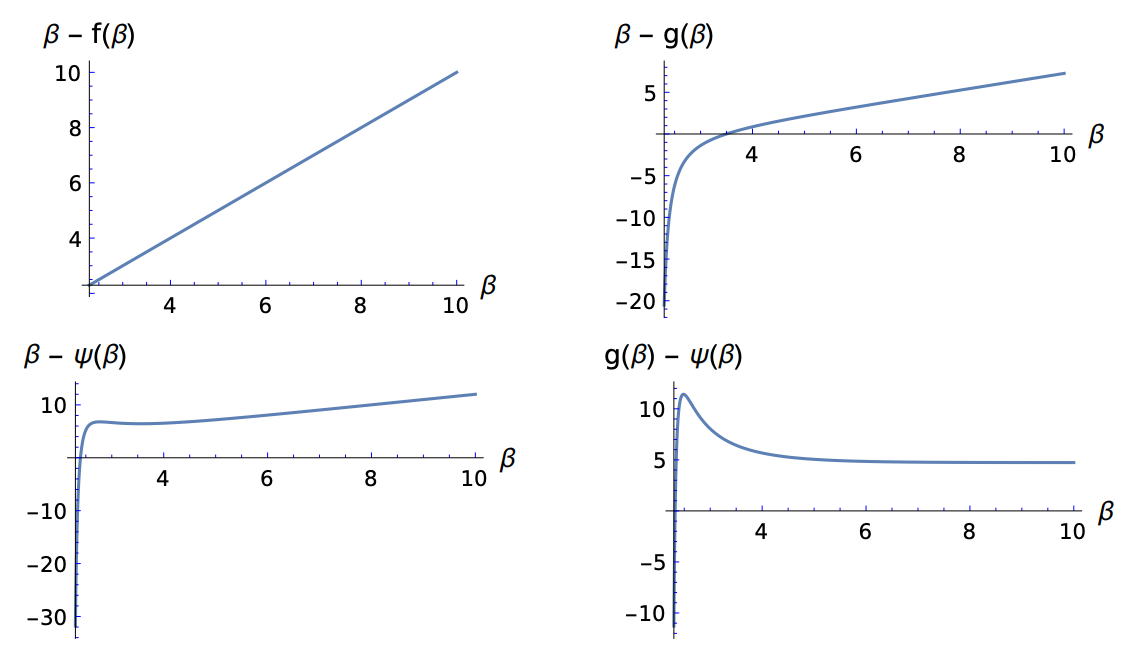}
\caption{Constraints on $\betaz$. The second top panel shows that only values $\betaz>3.5$ are 
admissible in the solutions to the stationary points of the loss in cases where the inputs are misclassified.}
\label{betaConstraints}
\end{figure}

From the figure, we see that the only active constraint is on logits that misclassified examples, where 
we see that stationary solutions require $\betaz>3.5$. In a surprising twist, it turns out that, empirically, 
adversarially robust logits happen to have a distribution with a mean that is close to 3.5! 

To summarize our findings thus far, we have found that for each training example, 
\tabL
\item[-]
we can explicitly determine logits that are stationary points of the cross-entropy loss 
--~equations (\ref{logitsCase1}) \& (\ref{logitsCase2})~-- depending 
on whether the example was misclassified or not.
\item[-]
We obtain a pair of solutions per logit, depending on the sign of the square root terms in 
(\ref{logitsCase1}, \ref{logitsCase2}).
\item[-]
The critical logits for misclassified examples have their maximum constrained by the function $\gPerp_\pm$ in (\ref{logitsCase2}) as seen in figure (\ref{betaConstraints}).
\tabR

With the solutions to the stationary points (\ref{logitsCase1}, \ref{logitsCase2}) in hand, we can establish their utility by substituting the solutions back into the expression for the {\em exact} cross-entropy loss (as opposed to the approximate loss used in our derivation), and examine their plots.
For ease of exposition, we confine ourselves to the positive root solutions.

For the purposes of visualizing the loss, we need to reduce the vast number of the $\{\betaz\}$ parameters determining the stationary points. There are of course $\Ndata$ such parameters, 
one for each sample in the training data. We propose lumping the values of $\betaz$ into two groups depending on whether $\logitvec(\vekx)$ led to a correct classification or not. 
Hence, we set 
$$
\betaz(\vekx)
=
\left\lbrace
\begin{array}{c l}
\betaCorrect & \mbox{if $\vekx$ is correctly classified
(occurs at a rate $1-\error$)
}
\\
\betaWrong & \mbox{ if $\vekx$ is misclassified 
(occurs at a rate $\error$)
}
\end{array}
\right.
$$

and take into account the constraint $\betaWrong>3.5$ as discussed above.

\begin{figure}[H]
\centering
\includegraphics[width=0.5\linewidth]{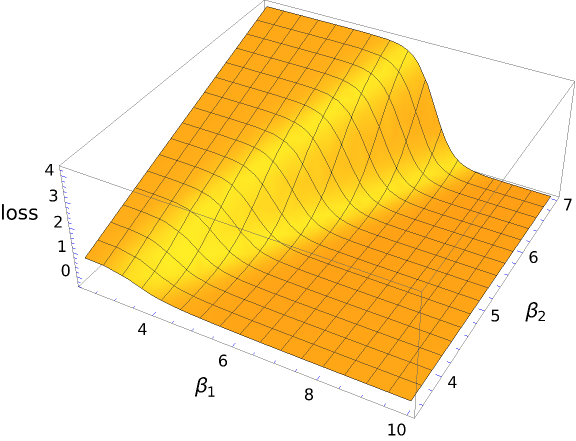}
\caption{The ``mean-field "loss surface as a function of $(\betaCorrect,\betaWrong)=(\beta_1, \beta_2)$}
\label{loss_surface}
\end{figure}

From figure (\ref{loss_surface}), we discern that
\begin{itemize}
\item[(a)]
large values of $\betaCorrect$ give rise to stationary points corresponding to 
near zero values of the loss. On the other hand, as seen in the top left panel of figure (\ref{betaConstraints}), large values of  $\betaCorrect, \betaWrong$ correspond to large values of the logit gaps $\betaCorrect-\fPerp( \betaCorrect)$ and  $\betaWrong-\gPerp( \betaWrong)$. Hence, within this framework, large logit gaps give rise to stationary points with low loss values. 
\item[(b)]
Small gap values for ``incorrect'' logits paired with small-to-moderate logit gaps for the 
``correct'' logits also yield low loss values. 
\item[(c)]
Large gap values for ``incorrect'' logits paired with small logit gaps for the 
``correct'' logits leads to higher loss values. 

\end{itemize}

Our central concern is understanding how the conclusions in items (a) and (b) above are modified when we include adversarial perturbations in our setup. 


\subsubsection{Adversarial Perturbations}
Now that we have a handle on some expected features of the logits of a model that achieves a given 
accuracy on the training data,  we can ask what happens to these logits when the model is 
subjected to adversarial perturbations. 

For analytic tractability, we confine ourselves to the simplest version of the so-called FGSM adversarial attacks \cite{harnessing} where each input feature vector $\vekx$ undergoes a perturbation
$\vekx \mapsto \vekx +\epsilon\delta\vekx$, with 
\bea
\delta \vekx := 
\frac{1}{\lVert\nabla_{\smallerB\vekx}\CE{\vek{\lmsd{z}}(\vekx)}{\veky}\rVert}
\nabla_{\smallerB{\vekx}}\CE{\logitvec(\vekx)}{\veky}
=
\frac{1}{\lVert\nabla_{\smallerB\vekx}\CE{\vek{\lmsd{z}}(\vekx)}{\veky}\rVert}
\slotJac{}\dg
\nabla_{\smallerB\logitvec}\CE{\logitvec}{\veky}
\label{attack}
\eea

where $\slotJac{}$ is the Jacobian of the map 
$\vekx\to\logitvec(\vekx)$. The perturbation in (\ref{attack}) then induces a perturbation 
to the logits $\logitvec(\vekx)\mapsto
\logitvec(\vekx+\epsilon\delta\vekx):=\logitvec(\vekx)+\delta\logitvec(\vekx)$, where

\bea
\delta\logitvec(\vekx) 
&=&\epsilon \slotJac{}\delta\vekx + \Oh(\epsilon^2)
\nonumber\\
&=&
\frac{\epsilon }{\lVert\nabla_{\smallerB\vekx}\CE{\vek{\lmsd{z}}(\vekx)}{\veky}\rVert}
\slotJac{}\slotJac{}\dg
\nabla_{\smallerB\logitvec}\CE{\logitvec}{\veky}
 + \Oh(\epsilon^2)
\label{logitAttack}
\eea

The $\Oh(\epsilon^2)$ truncation above assumes we are working in the {\em linear response} regime. 
A rigorous justification of the truncation would require demonstrating that the $\Oh(\epsilon^2)$ term 
has a strictly smaller norm than the linear contribution. We take this as granted for now.

Hence understanding of the effects of adversarial attacks on the logits of a trained neural net requires estimating the Jacobian of the neural net's input-output mapping. Obtaining an estimate of this Jacobian is a highly non-trivial problem and constitutes the most technical part of our analysis. In what follows, we present the results of the calculation, relegating the technical details of the derivation to later sections.

We are ultimately interested in figuring out the effects of the perturbation on the optimal logits defined  
in (\ref{logitsCase1}, \ref{logitsCase2}). 

Let $1\le \mu,\nu \le \Ndata$ denote sample indices for the data used to train the model, such that 
$\logitvec^{\mu}$ denotes the logit output given training input $\vekx^{\mu}$. Similarly, 
$\veky^{\mu}$ denotes the ground truth label for the input $\vekx^{\mu}$. 

We will show that under an FGSM attack, equation (\ref{logitAttack}) takes the form

\bea
\delta\logitvec^\mu =
\epsilon
\sum_{\nu=1}^{\Ndata}
\Omega_{\nu}
\left[
\logitvec^{\nu}\cdot(\vek{1}-\Nclasses\veky^\mu)
\right]
\logitvec^{\nu} +  \Oh(\epsilon^2)
\label{logitShift}
\eea

where the coefficients $\Omega_{\nu}$  are non-negative functions of the population input features $\{\vekx^\mu\}$ in the  training data, and otherwise play only a peripheral role in our discussion. 

We would like to understand how the logit gaps change under adversarial perturbations. To do so, we 
substitute our surrogate logits from equations (\ref{logitsCase1}, \ref{logitsCase2}) into 
equation (\ref{logitShift}), and consider examples $\vekx^\mu$ which are {\em correctly classified in the absence of adversarial attacks}. 

As detailed in the previous sections, within our framework, correctly classified examples have  logit gaps 

\[
\max\{\logitvec^\mu\}-
2^{\mathrm{nd}}\textrm{-}\max\{\logitvec^\mu\}
=
\beta_{\smaller{\logitvec^\mu}}-\fPerp(\beta_{\smaller{\logitvec^\mu}})
\,\,\, \leftrightarrow \,\,\, \betaCorrect - \fPerp(\betaCorrect)
\]

which, when directly substituted into (\ref{logitShift}) gives

\be
\delta\Big(
\betaCorrect - 
\fPerp(\betaCorrect)
\Big)
=
\frac{\epsilon}{\Nclasses-1}
\sum_{\nu=1}^{\Ndata}
\Omega_{\nu}
\big(
(\logitvec^{\nu}\cdot\vek{1})^2
-\Nclasses
\logitvec^{\nu}\cdot \logitvec^{\nu}
\big)
+
\Oh(\epsilon^2)
\label{deltaGap}
\ee

Equation (\ref{deltaGap}) provides an explicit expression for the expected change in the 
logit gaps due to adversarial perturbations. More precisely, we first partition the sum over the training data on the RHS of  (\ref{deltaGap}) into disjoint subsets consisting of correctly classified and misclassified examples as 
$\displaystyle{
\{1\le\nu\le\Ndata\}=
\{{\nu \in \correct}\}\cup
\{{\nu \in \wrong}\}.
}$

After setting 

\[
(1-\error)\Omega_{\correct}(\epsilon) := 
\sum_{\nu\in\correct}\epsilon\Omega_\nu
\qquad\mbox{and}\qquad
\error\,\Omega_{\wrong}(\epsilon) := 
\sum_{\nu\in\wrong}\epsilon\Omega_\nu
\]

 we substitute our surrogate logit solutions (\ref{logitsCase1}, \ref{logitsCase2}) into 
 the RHS of (\ref{deltaGap}) to obtain

\bea
&&
\delta\Big(
\betaCorrect - 
\fPerp(\betaCorrect)
\Big)
\nonumber \\
&=&
-
\left[
(1-\error)\Omega_{\correct}(\epsilon)
\Big(
\betaCorrect - 
\fPerp(\betaCorrect)
\Big)^2
+
\error\Omega_{\wrong}(\epsilon)
\Big(
\betaWrong - 
\gPerp(\betaWrong)
\Big)^2
\right]
\nonumber \\
&&\qquad
-
2\frac{\Nclasses-2}{\Nclasses-1}
\error\Omega_{\wrong}(\epsilon)
\Big(\betaWrong-\gPerp(\betaWrong)\Big)
\Big(\gPerp(\betaWrong) - \psi_\wrong\Big)
\nonumber \\
\label{deltaGap2}
\eea

Equation (\ref{deltaGap2}) provides the following insights.

\begin{itemize}
\item
From figure (\ref{betaConstraints}), we know that for $\betaWrong>3.5$, the quantities appearing 
in (\ref{deltaGap2}) are all positive. Hence adversarial attacks have the effect of shrinking the logit gaps of logits which correctly classify examples prior to the attack.

\begin{siderules}
Intuitively, one would expect that logits 
with small gaps will be closer to the classification error threshold (as these are more prone to having their class predictions flipped). It therefore makes sense that the primary effect of an adversarial attack is to close the logit gap as this is a surefire way of pushing a model's logits towards the error threshold.
\end{siderules}

\item
The magnitude of the ``gap shrinkage''  $\delta\big(\betaCorrect - \fPerp(\betaCorrect)\big)$ due to adversarial attacks is dependent on the model's classification error rate $\error$ prior to the attack.
Hence, we predict that models that overfit the training data (i.e. $\error\downarrow 0)$ will fail to 
be robust to adversarial attacks if they exhibit large logit gaps when trained on unperturbed data. 
\item
The magnitude of the change in the logit gap due to adversarial perturbations is quadratically dependent on the logit gaps 
$\betaCorrect - \fPerp(\betaCorrect)$ and $\betaWrong - \fPerp(\betaWrong)$ obtained 
prior to the adversarial attack.
\begin{siderules}
It follows that logits with logit gaps $\gg1$ are much more susceptible to large  perturbations in their response to adversarial attacks relative to logits with gaps $\ll 1$.
\end{siderules} 
\end{itemize}

\begin{figure}[H]
\centering
\hspace*{-1.5cm}
\includegraphics[width=1.35\linewidth]{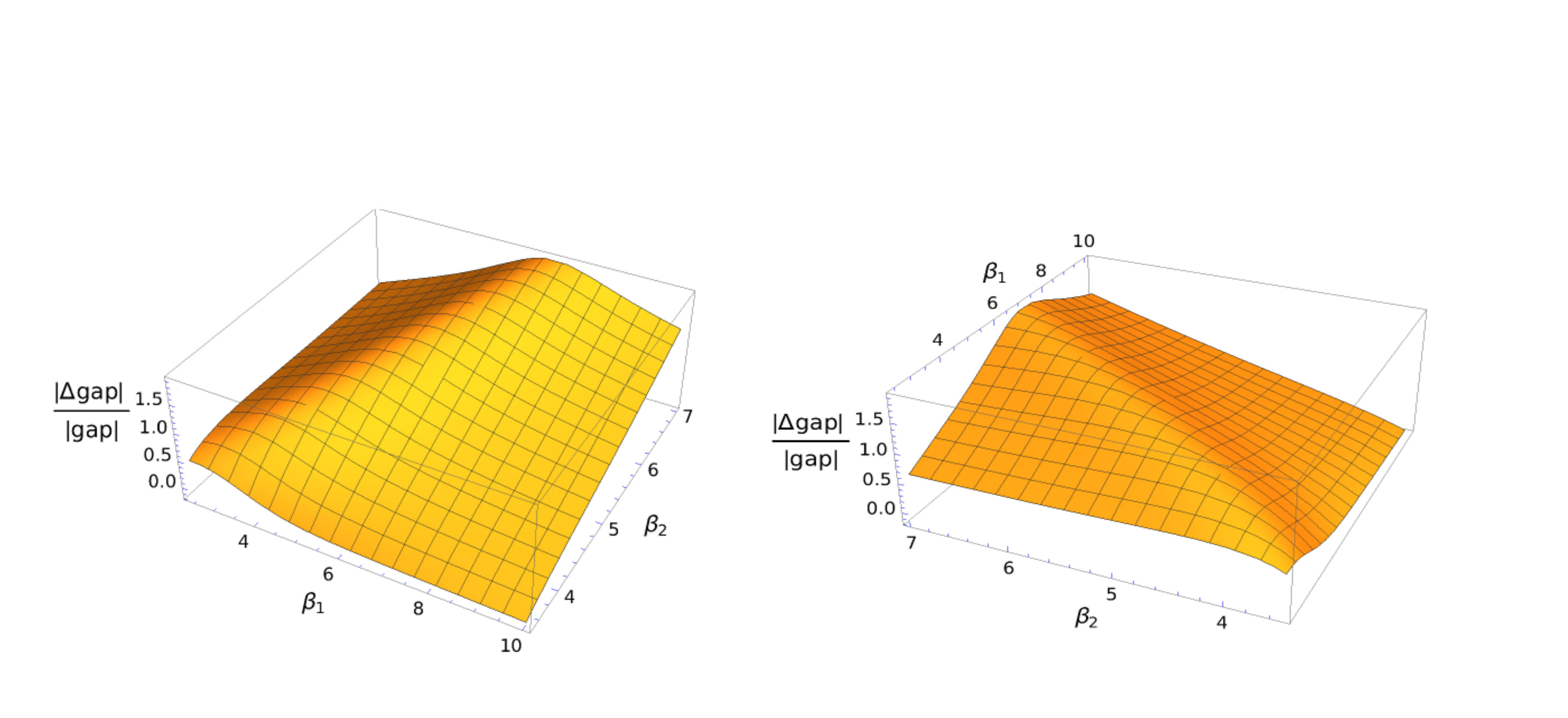}
\caption{Content of equation (\ref{deltaGap2}) seen from two vantage points emphasizing the relative change in the logit gap due to adversarial perturbations as a function of $(\betaCorrect,\betaWrong)=(\beta_1, \beta_2)$ with $\epsilon_{\mathrm{adversarial}}=0.1$ and unperturbed input vectors normalized to unit length. A model is said to be robust if $\displaystyle{\frac{\vert\Delta\mathrm{gap}\vert}{\vert\mathrm{gap}\vert}}\ll \Oh(\epsilon_{\mathrm{adversarial}})$.}
\label{deltaGapFig1}
\end{figure}

The content of equation (\ref{deltaGap2}) is captured in Figure (\ref{deltaGapFig1}), showing how the logit gaps in equation (\ref{deltaGap2}) depend 
on the parameters $(\betaCorrect, \betaWrong)$.

\subsection{Estimating the Jacobian}
\subsubsection{Linear Response Regime}
The goal of this section is to calculate the change to the logits, $\delta\logitvec^\mu$, induced by 
an adversarial perturbation $\vekx^\mu\to\vekx^\mu+\epsilon\delta\vekx^\mu$, where 
$\epsilon$ is the strength of the perturbation. We will do this in 
the {\em linear response} regime where 

\be
\delta\logitvec^\mu = \slotJac{\mu}\epsilon\delta\vekx^\mu + \Oh(\epsilon^2)
\label{linearResponse}
\ee

and $\slotJac{\mu}$ is the jacobian of the deep neural net mapping $\vekx^\mu\mapsto\logitvec^\mu$ from inputs to logits. 
We confine our analysis to simple FGSM attacks \cite{harnessing} of the form

\[
\delta\vekx^\mu = 
\frac{\nabla_{\smallerB\vekx^\mu}\CE{\logitvec^\mu}{\veky^\mu}}
{\lVert
\nabla_{\smallerB\vekx^\mu}\CE{\logitvec^\mu}{\veky^\mu}
\rVert}
=
 \zeta_{\smaller\vekx^\mu, \veky^\mu, \logitvec^\mu}
 \slotJac{\mu}\dg\nabla_{\smallerB\logitvec^\mu}\CE{\logitvec^\mu}{\veky^\mu},
 \,\,
 \zeta_{\smaller\vekx^\mu, \veky^\mu, \logitvec^\mu}
 :=\frac{1}{\lVert
 \slotJac{\mu}\dg\nabla_{\smallerB\logitvec^\mu}\CE{\logitvec^\mu}{\veky^\mu}
 \rVert
 }
\]

Hence, under adversarial FGSM attacks, the logits exhibit a linear response 

\be
\delta\logitvec^\mu = 
\epsilon  \zeta_{\smaller\vekx^\mu, \veky^\mu, \logitvec^\mu}
\slotJac{\mu} \slotJac{\mu}\dg\nabla_{\smallerB\logitvec^\mu}\CE{\logitvec^\mu}{\veky^\mu} + \Oh(\epsilon^2)
\label{linearResponse}
\ee

\subsubsection{Random Matrix Formulation}
From a technical viewpoint, it is convenient to formulate the problem in terms of random matrices. We pause to introduce the relevant notation. 

\begin{itemize}
\item
In what follows,  $\langle f \rangle$ denotes the average of $f$ relative to a centered, unit-variance Gaussian distribution.
\item As before, our training data consists of input feature vectors $\{\vekx^\mu\}_{\mu=1}^{\Ndata}$,  with each $\vekx^\mu\in\R^{\Nfeats}$, with $\Nfeats$ denoting their dimensionality.
\item 
Let $\Xx$ denote the $\Ndata\times\Nfeats$ matrix whose rows are comprised of the input feature vectors $\{\vekx^\mu\}_{\mu=1}^{\Ndata}$.
\item 
Similarly, $\Zz$ denotes the $\Ndata\times\Nclasses$ matrix whose rows are comprised of the output logit vectors $\{\logitvec^\mu\}_{\mu=1}^{\Ndata}$.
\end{itemize}

In this formulation, a deep neural net is trained to map $\Xx\to\Zz$, and our surrogate logits in (\ref{logitsCase1}, \ref{logitsCase2}) provide us with estimates $\Zzt$ which depend on $\Xx$ implicitly through the distribution 
of the maximum values of the rows of $\Zz$. Our goal is to make this dependence explicit to facilitate computing 
the Jacobian of the mapping. 

To do so, we pass through an auxiliary problem by constructing a matrix-valued mapping $
\upomega: (\Xx, \Zzt) \mapsto \upomega(\Xx, \Zzt)$, such that 
\begin{itemize}
\item
$\upomega(\Xx, \Zzt)$ is an $\Nfeats\times\Nclasses$ matrix
\item
$\Xx\upomega(\Xx, \Zzt) - \Zzt$ is a random $\Ndata\times\Nclasses$ matrix with {\em iid} centered Gaussian entries 
with variance $\sigma_0^2$. 
\end{itemize}

Note that this imposes a constraint on the expectation value of the trace: 

\[
\Tr
\left
\langle
\upomega(\Xx, \Zzt)\upomega(\Xx, \Zzt)\dg\Xx\dg\Xx
\right
\rangle
=
\Tr(\Zzt\dg\Zzt)+\sigma_0^2\Ndata\Nclasses
\] 

which implies that any feasible solution $\upomega(\Xx,\Zzt)$ must have a norm

\[
\lVert\upomega(\Xx, \Zzt) \rVert
\sim
\frac{1}{\lVert\Xx \rVert}
\sqrt{ \lVert\Zzt \rVert^2 +  \sigma_0^2\Ndata\Nclasses}
\]

The idea behind this construction is that, as $\sigma_0^2\to 0$, the matrix $\Xx\upomega(\Xx, \Zzt)$ is forced to be 
arbitrarily close to the surrogate logit matrix $\Zzt$. The motivation being that computing the Jacobian of the map
 $\Xx \to \Xx\upomega(\Xx, \Zzt)$ is relatively straightforward. 

Without loss of generality, we normalize our input feature vectors to unit length so that $\lVert\Xx\rVert=\sqrt{\Ndata}$. 
Taking into account the norms of the surrogate logits in (\ref{logitsCase1}) suggests that $\lVert\upomega(\Xx, \Zzt) \rVert$ 
should be bounded by $c\sqrt{\Nfeats\Nclasses}$, with $c$ a positive constant of $\Oh(1)$.  

This makes our optimization problem tractable as we can look for solutions that lie on a convex 
subspace of $\Nfeats\times\Nclasses$ matrices. The solution to this problem 
is given in a beautiful piece of analysis by Fyodorov in \cite{fyodorov2019}, by minimizing 
the energy functional 

\[
\scr{E}(\Xx, \Zzt, \sigma_0\mat{W}, \lambda):=
\lVert
\Xx\upomega(\Xx, \Zzt)-
\Zzt-\sigma_0\mat{W}
\rVert^2
-
\lambda\lVert\upomega(\Xx,\Zzt) \rVert^2
\]

where, as above, $\mat{W}$ is a random matrix with entries drawn from a centered, unit-variance Gaussian distribution, 
and $\lambda$ is a Lagrange multiplier that enforces the constraint on the norm of $\upomega(\Xx, \Zzt)$.

Fyodorov's solution yields 

\be
\upomega(\Xx,\Zzt) = \big[\Xx\dg\Xx-\lambda_\star\mat{1}\big]^{-1}\Xx\dg\big(\Zzt-\sigma_0\mat{W}\big)
\label{fyodorov}
\ee

with the Lagrange multiplier $\lambda_\star$ as the solution to 

\[
c^2\Nfeats\Nclasses = 
\Tr
\left(
\Xx\big[\Xx\dg\Xx-\lambda_\star\mat{1}\big]^{-2}\Xx\dg\big(\Zzt\Zzt\dg+\sigma_0^2\mat{1}\big)
\right)
\]

For a complete account of the solution, we refer the reader to \cite{fyodorov2019}.

\subsubsection{The Jacobian}
We now have all the pieces needed to compute the Jacobian. The previous section provides us 
with the estimate 

\[
\Zz = \sigma_0\mat{W} + \Xx\big[\Xx\dg\Xx-\lambda_\star\mat{1}\big]^{-1}\Xx\dg\big(\Zzt-\sigma_0\mat{W}\big)
\]

or equivalently, with $\vek{w}^{\mu}, \vekx^\mu, \logitvec^\mu$ denoting the $\mu^{\mathrm{th}}$ row of 
$\mat{W}, \Xx, \Zz$ respectively, 

\[
\logitvec^\mu = \sigma_0\vek{w}^{\mu}+
\big(\Zzt-\sigma_0\mat{W}\big)\dg\Xx
\big[\Xx\dg\Xx-\lambda_\star\mat{1}\big]^{-1}
 \vekx^\mu.
\]

As shown in  \cite{fyodorov2019}, 
the Lagrange multiplier $\lambda_\star$ approaches a fixed constant as $\sigma_0\to0$.  In this case, a straightforward but lengthy computation of the Jacobian of the mapping $\vekx^\mu\to\logitvec^\mu$ yields 

\beaNN
\slotJac{\mu}_{m j}
&=&
\sum_{\nu=1}^{\Ndata} \tilde{z}^{\nu}_m
\frac{\partial}{\partial x^{\mu}_{j}}
\Big(\vekx^\nu\cdot[\Xx\dg\Xx-\lambda_\star\mat{1}]^{-1}\vekx^\mu\Big)
\\
&=&
\Big(
[\Xx\dg\Xx-\lambda_\star\mat{1}]^{-1}\Xx\dg
\Big)_{j\mu}
\tilde{z}^{\mu}_m
+
\left[
\Big(
[\Xx\dg\Xx-\lambda_\star\mat{1}]^{-1}\Xx\dg
\Big)\Zzt
\right]_{jm}
\eeaNN

so that the matrix $\slotJac{\mu}\slotJac{\mu}\dg$ appearing in the linear response 
formula (\ref{linearResponse}) has entries 

\beaNN
\left(
\slotJac{\mu}\slotJac{\mu}\dg
\right)_{mn}
&=&
\Big[
\Xx
[\Xx\dg\Xx-\lambda_\star\mat{1}]^{-2}\Xx\dg
\Big]_{\mu\mu}
\tilde{z}^{\mu}_m
\tilde{z}^{\mu}_n
+
\Big[
\Zz\dg\Xx
[\Xx\dg\Xx-\lambda_\star\mat{1}]^{-2}\Xx\dg
\Zzt
\Big]_{mn}
\\
&&\qquad
+\qquad
\tilde{z}^{\mu}_m
\Big[
\Xx
[\Xx\dg\Xx-\lambda_\star\mat{1}]^{-2}\Xx\dg
\Zzt
\Big]_{\mu n}
+
\tilde{z}^{\mu}_n
\Big[
\Xx
[\Xx\dg\Xx-\lambda_\star\mat{1}]^{-2}\Xx\dg
\Zzt
\Big]_{\mu m}
\eeaNN

Finally, setting 

\[
\Omega(\Xx,\lambda_\star)
:= 
\Xx
[\Xx\dg\Xx-\lambda_\star\mat{1}]^{-2}\Xx\dg
\]

leads to the expression (\ref{logitShift}) for the response of the logits under FGSM attacks.

\end{document}